\documentclass[11pt]{article}

\usepackage[preprint]{acl}

\usepackage{times}
\usepackage{latexsym}
\usepackage{url}

\usepackage[T1]{fontenc}

\usepackage[utf8]{inputenc}

\usepackage{microtype}

\usepackage{inconsolata}

\usepackage{graphicx}

\usepackage{booktabs}   
\usepackage{multirow}   
\usepackage{adjustbox}  
\usepackage{tabularx}
\usepackage{makecell}
\usepackage{enumitem}
\usepackage{array}
\usepackage{ragged2e}
\usepackage{algorithm}
\usepackage{algorithmic}
\usepackage{amsmath}
\title{EmoTrace: An Emotion Trajectory-Centered Framework for Psychological Support Dialogue Generation}

\author{
Kaitong Weng\textsuperscript{1},~ 
Lixin Liu\textsuperscript{1},~
Zihao Liu\textsuperscript{1},~
Bo Wang\textsuperscript{2},*,~
Shiguang Ni\textsuperscript{1},*
\\
\textsuperscript{1}Shenzhen International Graduate School, Tsinghua University\\
\textsuperscript{2}Department of Psychological and Cognitive Sciences, Tsinghua University
\\[0.5em] 
\texttt{bo-wang@tsinghua.edu.cn, ni.shiguang@sz.tsinghua.edu.cn}
}

\begin{document}
\maketitle

\begin{abstract}
Using large language models (LLMs) to assist psychological counseling is an important task in the field of natural language processing. The construction of high-quality psychological support dialogue corpora serves as a critical foundation for training counseling-oriented conversational models. However, existing data generation approaches generally suffer from several limitations, including emotionally stable seekers, limited variation in emotional dynamics, and a high degree of compliance with counselors’ guidance. These issues result in LLM that lack the capability to effectively respond to emotionally unstable scenarios. In addition, counselor responses are typically driven by problem-solving objectives, thereby overlooking the role of emotion-focused interaction, which are essential in psychological counseling. To address these gaps, we propose \textbf{EmoTrace}, a multi-turn dialogue corpus generation framework centered on modeling seekers’ emotional trajectories. we construct seekers' cognitive profile and introduce a seeker module with emotional schemas and an associated activation mechanism, a counselor module, and an emotional trajectory control module, thereby enhancing the layering of the seeker's emotional expression and the counselor's targeted empathic expression. Experimental results demonstrate that the proposed method outperforms existing approaches in terms of emotional richness and empathy quality. The complete dataset and the fine-tuned model will be made publicly available once the paper is accepted.
\end{abstract}

\renewcommand{\thefootnote}{}
\footnotetext{\(^{*}\) Corresponding author.}
\renewcommand{\thefootnote}{\arabic{footnote}}

\section{Introduction}

Mental health constitutes a fundamental component of overall health, with approximately 14\% of the global disease burden attributable to neuropsychiatric disorders~\citep{prince2007mental}. However, due to its longstanding underrepresentation in global health priorities, mental health services continue to face stigma and limited accessibility. As a result, many affected individuals are unable to obtain timely care~\citep{truzoli2026mental}.

The rapid development of large language models (LLMs) has introduced new perspectives for psychological counseling applications. Models such as GPT~\citep{openai2023gpt4} and LLaMA~\citep{touvron2023llama} have demonstrated human-like emotional support capabilities~\citep{sorin2024large}. Building on this progress, a number of studies have proposed specialized models for the counseling domain, including MeChat~\citep{qiu2024smile} and CpsyCounX~\citep{zhang2024cpsycoun}. These models rely on real-world or synthetic dialogue corpora to learn general patterns of psychological counseling. Consequently, the construction of high-quality dialogue corpora becomes a critical factor influencing model performance~\citep{qi2025kokorochat}. However, due to constraints related to medical ethics and personal privacy protection, the construction of large-scale, compliant mental health support dialogue datasets remain a central challenge in developing LLM-based counseling systems~\citep{mayer2022gdpr}.

\begin{figure*}[htbp]
    \centering
    \includegraphics[width=\textwidth]{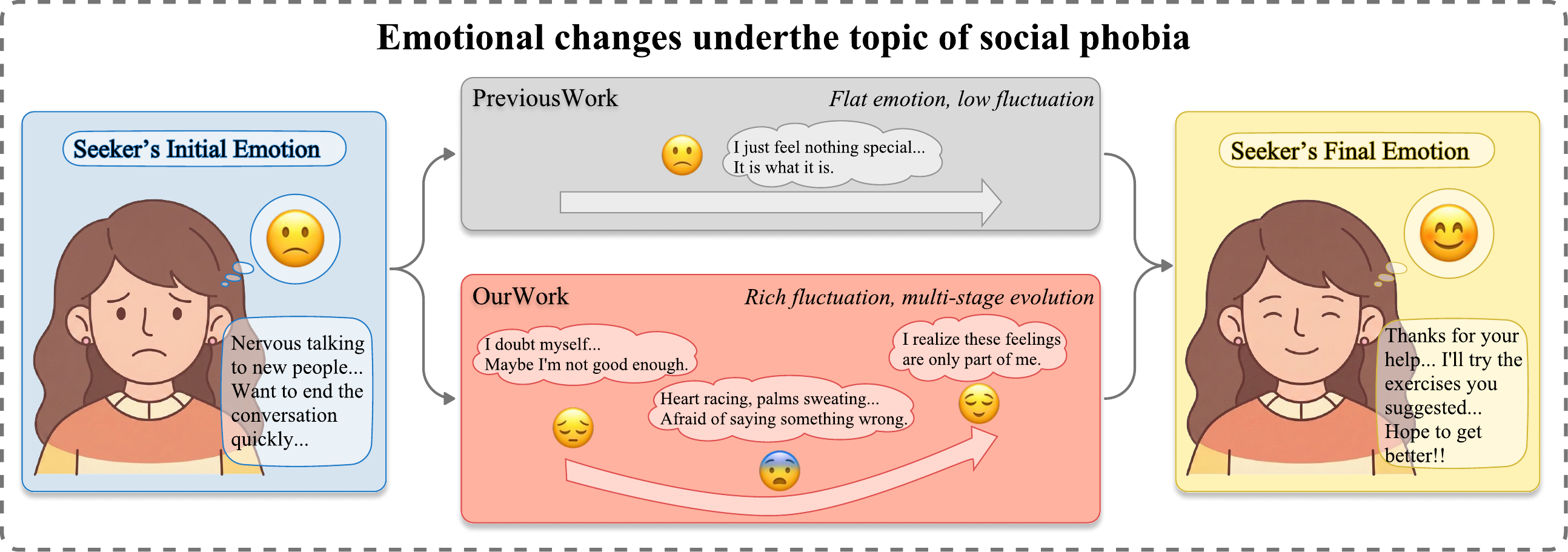}
    \caption{A comparison of the emotion trajectories of seekers in traditional datasets and EmoTrace-D on the topic of social phobia.}
    \label{fig:Trajectory_Comparison}
\end{figure*}

Recently, several studies on automated corpus construction, such as DeepWell-Adol~\citep{qiu2025deepwell} and PsyDT~\citep{xie2025psydt}, have focused on modeling and optimizing the counselor side of the psychological support dialogue. These approaches improve response quality by simulating counselors’ linguistic styles and therapeutic orientations or by employing dialogue frameworks grounded in positive psychology. 

However, this counselor-centered paradigm for corpus construction exhibits some limitations. As shown in \autoref{fig:Trajectory_Comparison}, the seekers’ emotional states in conventional dialogue often show low variability and a homogeneous progression, accompanied by a high level of compliance. Such patterns fail to reflect the emotional fluctuations commonly observed in real-world interactions. Training corpora constructed by this paradigm make it difficult for models to learn response strategies in the face of complex emotional dynamics. Furthermore, existing work typically characterizes seekers using emotion labels~\citep{liu2021towards} and the Big Five personality traits~\citep{yang2025psyplay}. However, they hardly considered the underlying cognitive structures of emotions, such as \textit{core beliefs} (individuals’ deepest self-perceptions and value systems)~\citep{beck1979ctd} and \textit{emotional schemas} (cognitive evaluation and coping patterns of emotional experiences)~\citep{leahy2002emotional}. This approach limits seekers to narrating only events, lacking cognitive evaluations of core beliefs or emotional schemas, so the model struggles to capture the logical chain from event to cognition to emotion.

Additionally, prior corpus construction work also have certain limitations in the design of counselor response strategies. Within the humanistic therapy tradition, Emotion-Focused Therapy (EFT) represents a quintessential approach~\citep{greenberg2004eft}. EFT emphasizes the acceptance and integration of emotional experience, adhering to a seeker-centered framework. However, in existing work, counselor responses are largely oriented toward direct problem-solving, while emotional processing is confined to surface-level empathy. This approach neglects the central role of emotion in the process of psychological support~\citep{elliott2010eft}.

\begin{figure*}[htbp]
    \centering
    \includegraphics[width=\textwidth]{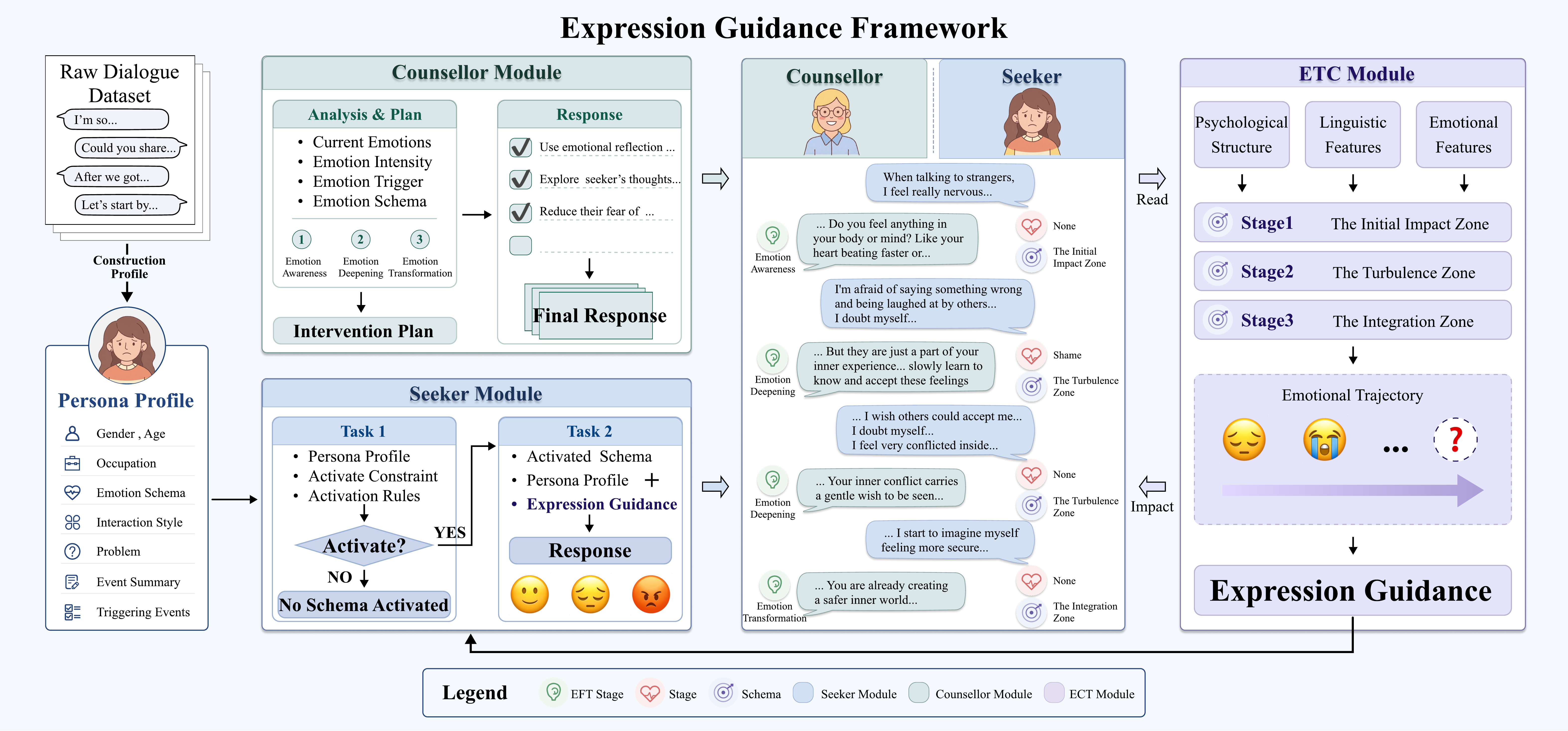}
    \caption{ The overall framework of EmoTrace.}
    \label{fig:framework}
\end{figure*}

In this paper, we propose \textbf{EmoTrace}, a seeker \textbf{Emo}tion \textbf{Tra}jectory-\textbf{Ce}ntered framework for multi-turn psychological support dialogue generation. Our method employs interactive role-playing to simulate dialogues between a seeker and a counselor. Specifically, in the seeker profile construction stage, we add basic personal information with an emotional schema dimension. The corpus generation is accomplished through the collaboration of three core modules. The overall framework of EmoTrace is illustrated in \autoref{fig:framework}. \textbf{The seeker module} generates responses that maintain personality consistency by integrating the persona profile with an emotional schema activation mechanism, while applying frequency constraints to avoid excessive schema activation. \textbf{The counselor module}, grounded in EFT, produces more targeted intervention strategies and responses. Finally, \textbf{the Emotion Trajectory (ETC) Control module} guides the seeker’s emotional development using a three-stage structure, modeling emotion as a trend variable to achieve controllable trajectories without compromising naturalness. 

Based on this method, we construct \textbf{EmoTrace-D}, a psychological support dataset containing 1,114 multi-turn dialogues. We also fine-tune an open-source LLM named \textbf{EmoTrace-M} on EmoTrace-D. Extensive experimental results demonstrate that, compared with prior work, the proposed method significantly improves the emotional reality and complexity of the generated corpora and boosts the empathic alignment of the model when using the data. Our contributions are as follows:

\begin{itemize}
    \item We propose \textbf{EmoTrace}, the first paradigm shifting corpus construction from counselor-centered to seeker emotional trajectory-centered, enabling seekers as dynamic emotional entities for more authentic dialogues.
    \item We construct the \textbf{EmoTrace-D} and fine-tune the \textbf{EmoTrace-M} on this dataset. Extensive empirical experiments demonstrate that the model achieves superior targeted empathic capability compared to prior work.
\end{itemize}

\section{Related Work}
\subsection{Psychological Support Dialogue Datasets}

Early psychological support dialogue datasets primarily relied on crowdsourcing and online psychological counseling platforms as data sources, with support strategies annotated manually~\citep{welivita2022curating,sun2021psyqa,wang2019persuasion}. However, due to the highly sensitive nature of psychological counseling, the acquisition of relevant corpora is often subject to privacy restrictions. Consequently, the majority of existing research has adopted synthetic data approaches to construct corpora. For instance, SmileChat~\citep{qiu2024smile} extends PsyQA from single-turn question-answering to multi-turn dialogues using ChatGPT, while CpsyCounD~\citep{zhang2024cpsycoun} employs a two-stage framework based on counseling reports to generate 3,134 multi-turn dialogues. 

Recently, to enhance the professionalism of psychological support dialogue corpora, some studies integrate psychological theories into the dialogue generation process~\citep{lee2024cactus,zhou2025diacbt,zhou2025crisp,shi2026simulating}. However, these methods primarily focus on the selection and optimization of counseling strategies, without the modeling of the seeker's cognitive structure.

\subsection{Psychological Counseling Dialogue Systems}

The rapid development of LLMs has advanced the growth of AI-driven psychological counseling services. Recent studies, such as MindChat~\citep{xue2023mindchat} and SoulChat~\citep{chen2023soulchat}, apply supervised fine-tuning to foundation models using high-quality dialogue data, with a focus on improving empathetic capabilities. 

Given that psychological counseling involves tasks such as intervention strategy planning and empathetic response generation, it naturally aligns with the collaborative paradigm of multi-agent systems, leading to the growing adoption of multi-agent approaches for psychological counseling. For example, WiseMind~\citep{wu2026wisemind} employs two cooperative agents representing rational reasoning and emotional reasoning, together with a DSM-5 knowledge graph for psychiatric assessment, thereby improving diagnostic accuracy and reducing hallucinations. To address difficulties in cross-session memory retrieval, TheraMind~\citep{hu2026theramind} adopts a dual-loop agent architecture designed for long-term counseling, with capabilities for strategy planning and adaptive adjustment. However, these methods lack reasoning about the causes of the seeker's emotional changes, making it difficult to handle complex emotional scenarios.

\section{Methodology}

\subsection{Overview of the Psychological Support Dialogue Generation Framework}

We synthesize psychological support dialogues through interactive role-playing and core component control, including three modules: a seeker module, a counselor module, and an ETC module.

Firstly, based on the persona profile, the seeker module generates responses by integrating the emotional expression guidance provided by the ETC module with the activation status of emotional schemas. Then, the counselor module generates responses via a pipeline consisting of two submodules: the analysis-planning submodule formulates psychological intervention strategies by emotion analysis and schema recognition, while the generation submodule then produces responses based on the intervention strategy derived from the former. After each turn of interaction between the seeker and the counselor, the ETC module determines the seeker's current stage based on the dialogue history. It then infers the emotional dynamics for the next turn, and provides guidance for subsequent emotional expression.

\subsection{Design of Psychological Support Dialogue Generation}

\subsubsection{Preparation: Seeker Persona Profile Construction}

To enhance the authenticity and consistency of the seeker's expression patterns in psychological support dialogue corpora, we draw inspiration from personalized role-driven study~\citep{wang2024rolellm}, constructing persona profiles by extracting individual characteristics and psychological attributes from raw dialogue data. This enables the LLM-driven seeker to maintain consistent cognitive patterns and emotional response styles across multi-turn dialogues. We used 5,000 multi-turn dialogues from PsyDT as seed data and employed GPT-4.1-mini to summarize the seeker's background information and personality traits in each dialogue, organizing them into structured persona profiles. Each profile contains information such as gender, age, occupation, interaction style and problems.

Traditional persona profile construction typically characterizes individuals merely with emotion labels or simple background information, which fails to capture their underlying psychological mechanisms. To address this limitation, we propose the emotional schema as a modeling dimension to characterize individual traits across three levels: cognitive, emotional, and behavioral. Specifically, based on the classification of core beliefs in Cognitive Behavioral Therapy (CBT)~\citep{beck1979ctd} and the emotional schema model~\citep{leahy2002emotional}, we establish a three-dimensional cognitive axis: \textbf{the self axis} (cognitive style of one's own emotions), \textbf{the others axis} (cognitive style of others and interpersonal relationships), and \textbf{the world/future axis} (cognitive style of the persistence and meaning of emotions). Each axis reflects how individuals respond to emotional experiences from different perspectives. On this basis, we select eight representative emotional schemas, each linked to typical cognitive tendencies along these axes. For example, people with a guilt schema tend to feel self-reproach when their own actions negatively affect others. Detailed information on emotional schemas is provided in the Appendix \ref{app:persona_profile}.

Finally, we clustered and filtered all persona profiles according to topics and schema types, resulting in 1,423 uniformly distributed persona profiles. The prompt for generating persona profiles is provided in the Appendix \ref{app:persona_profile}.

\subsubsection{Emotional Trajectory Control Module}

This module provides controllable guidance of the seeker's emotional development process across multi-turn dialogues based on three-stage emotional trajectory modeling. 

\textbf{Stage-based emotional trajectory.} Emotional change in psychological counseling is gradual, not instantaneous~\citep{stiles1990assimilation}. Therefore, observing emotional change from a trajectory perspective can better captures the continuity of the seeker's emotional state and the process of emotional evolution than single time point judgments. 

Inspired by the EFT theory ~\citep{greenberg2012emotions}, we divide the seeker's emotional trajectory into three stages: the initial impact zone, the turbulence zone, and the integration zone. This stage-based design, compared with the fixed node approaches, allows natural emotion fluctuations within broader bounds. In this way, emotional transitions no longer take the form of discrete label switching but instead constitute an evolution process with inherent logic. See Appendix \ref{app:dialogues_syn} for stage definitions. 

\textbf{Emotional trajectory control.} Guided by the stage model, this module determines the seeker's current stage by analyzing the psychological structure, linguistic features, and emotional characteristics of the seeker's utterances. In real-world psychological counseling scenarios, the seeker's emotional development depends not only on their own willingness to change but also, to a large extent, on the counselor's response style~\citep{ehrlich1979counselor}. Accordingly, this module further analyzes how the counselor's response may influence the seeker's subsequent reply and infers the next-turn dynamics of emotional change. Finally, based on the above analysis, the module outputs guidance for seeker's emotional expression in the next turn, guiding the direction and intensity of emotional development within flexible stage boundaries, without enforcing deterministic emotional expression outcomes.

\subsubsection{Seeker Module}

To avoid homogenized expression and unnatural dialogue caused by excessive activation of emotional schemas, we designed a schema activation mechanism based on conditional triggering and frequency constraints. 

Specifically, we maintain a long-term and short-term activation sequence for each type of emotional schema that every seeker possesses. When a particular schema has been activated twice consecutively or five times in total, its activation is prohibited for the current turn. Only when the schema is not subject to such restriction does the mechanism determine, based on specified conditions, whether the schema is permitted to be explicitly expressed in the current turn. The pseudocode for this mechanism is provided in the \autoref{alg:schema_constraint}.

After completing the above determination, the module generates responses that exhibit stable personality and dynamic emotions based on the persona profile by integrating the emotional expression guidance provided by the ETC module for the current turn, as well as the schema activation status.

\begin{table}[htbp]
    \centering
    \small
    \caption{Statistics of Different Datasets. NoT., LoS., LoC. respectively represent average number of turns, average length of seeker’s response and average length of counselor’s response.}
    \label{tab:dataset_stats}
    \begin{tabular}{lccc}
        \toprule
        \multirow{2}{*}{Dataset} & \multicolumn{3}{c}{Statistics} \\
        \cmidrule(lr){2-4}
        & NoT. &  LoS. &  LoC. \\
        \midrule
        SMILECHAT    & 10.4 & 26.1  & 28.9  \\
        CpsyCounD    & 8.7  & 30.4  & 49.7  \\
        DeepWell-Adol& 10.0 & 28.4  & 58.5  \\
        PsyDTCorpus  & 18.1 & 31.6  & 58.1  \\
        EmoTrace-D   & 12.2 & 34.6  & 60.3  \\
        \bottomrule
    \end{tabular}
\end{table}

\begin{table*}[htbp]
\centering
\caption{Evaluation results for five datasets across three evaluation frameworks. The best results are highlighted in \textbf{bold}, and the runner-up results are \underline{underlined}.}
\label{tab:dataset_eval}
\small
\begin{tabular}{llccccc}
\toprule
\multirow{2}{*}{Evaluation Matrix} & \multirow{2}{*}{Indicator} 
& \multicolumn{5}{c}{Datasets} \\
\cmidrule(lr){3-7}
& & CpsyCounD & SMILECHAT & DeepWell-Adol & PsyDTCorpus & EmoTrace-D \\
\midrule

\multirow{4}{*}{CpsyCoun}
& Comprehensiveness (0--2) & \underline{1.91} & 1.855 & 1.908 & \textbf{1.95} & 1.9 \\
& Professionalism (0--4) & 2.90 & 2.44 & 3.35 & \textbf{3.94} & \underline{3.67} \\
& Authenticity (0--3) & 2.21 & 2.30 & 2.68 & \underline{2.93} & \textbf{2.94} \\
& Safety (0--1) & 0.98 & 0.98 & \textbf{1} & \textbf{1} & \textbf{1} \\

\midrule
\multirow{5}{*}{PsyDT}
& Emotional Empathy (0--3) & 1.79 & 2.01 & 2.43 & \underline{2.8} & \textbf{2.85} \\
& Cognitive Empathy (0--3) & 2.24 & 2.39 & 2.64 & \underline{2.77} & \textbf{2.78} \\
& Conversation Strategy (0--3) & 1.97 & 1.98 & 2.47 & \textbf{2.84} & \underline{2.80} \\
& State and Attitude (0--3) & 2.32 & 2.31 & 2.7 & \underline{2.81} & \textbf{2.89} \\
& Safety (0--1) & 0.93 & 0.94 & 0.99 & \textbf{1} & \textbf{1} \\

\midrule
\multirow{5}{*}{EmoTrace-E}
& Emotional Changes (0--5) & 2.20 & 3.00 & 2.90 & \underline{4.21} & \textbf{4.74} \\
& \makecell[l]{Emotional Intensity and\\Complexity (0--4)} 
& 1.66 & 2.60 & 2.15 & \underline{3.03} & \textbf{3.56} \\
& Quality of Empathy (0--5) & 1.83 & 2.50 & 2.75 & \underline{4.11} & \textbf{4.49} \\
& Autonomy (0--4) & 1.71 & 2.48 & 2.31 & \underline{3.21} & \textbf{3.73} \\
& Motivation for Growth (0--2) & 1.28 & 1.57 & 1.50 & \underline{1.93} & \textbf{1.99} \\

\bottomrule
\end{tabular}
\end{table*}

\subsubsection{Counselor Module}

A pipeline is designed in counselor module to generate responses in the following sequence: emotion analysis, intervention planning, and response generation.

\textbf{Analysis and planning submodule.} Grounded in EFT, this submodule divides the therapeutic process into three stages, namely emotion awareness, emotion deepening, and emotion transformation. Based on the seeker's utterances, the submodule analyzes the emotion type, intensity, triggers of the emotion and potential underlying emotional schemas. On the basis of this analysis, the submodule selects an appropriate EFT therapeutic stage, formulates the current intervention goals and schema-loosening strategies. Finally, it integrates the above information to produce a detailed intervention strategy, providing a basis for subsequent response generation.

\textbf{Response generation submodule.} After receiving the analysis and planning results, this submodule is responsible for converting the structured intervention strategy into natural and coherent supportive language. During generation, the model strictly adheres to the prescribed intervention plan and performs no additional emotion analysis or strategy adjustment, thereby ensuring consistency between the response and the planned intervention.

\subsection{Dialogue Generation}

We use a role-playing approach to generate multi-turn psychological support dialogues by GPT-4.1-mini based on persona profiles. As shown in \autoref{fig:framework}, each turn involves interaction between seeker and counselor, while the ETC module generates guidance for the seeker's emotional expression in the next turn based on the output of the current turn. A complete dialogue is produced by iterating the above process. 

To ensure the completeness of emotional progression, we filter data by two criteria: full coverage of the seeker's three emotional stages and reasonable stage duration to prevent excessive persistence. Through this process, we construct EmoTrace-D with 1,114 multi-turn dialogues. The related prompts and a complete dialogue case are presented in the Appendix \ref{app:dialogues_syn}.

\section{General Dataset Quality Evaluation}

To validate the quality of EmoTrace-D, we compare it with four publicly available psychological support dialogue datasets: CpsyCounD~\citep{zhang2024cpsycoun}, SMILECHAT~\citep{qiu2024smile}, DeepWell-Adol~\citep{qiu2025deepwell}, and PsyDTCorpus~\citep{xie2025psydt}. \autoref{tab:dataset_stats} presents the statistical information of these five datasets. 

\begin{figure}[htbp]
    \centering
    \includegraphics[width=0.85\columnwidth]{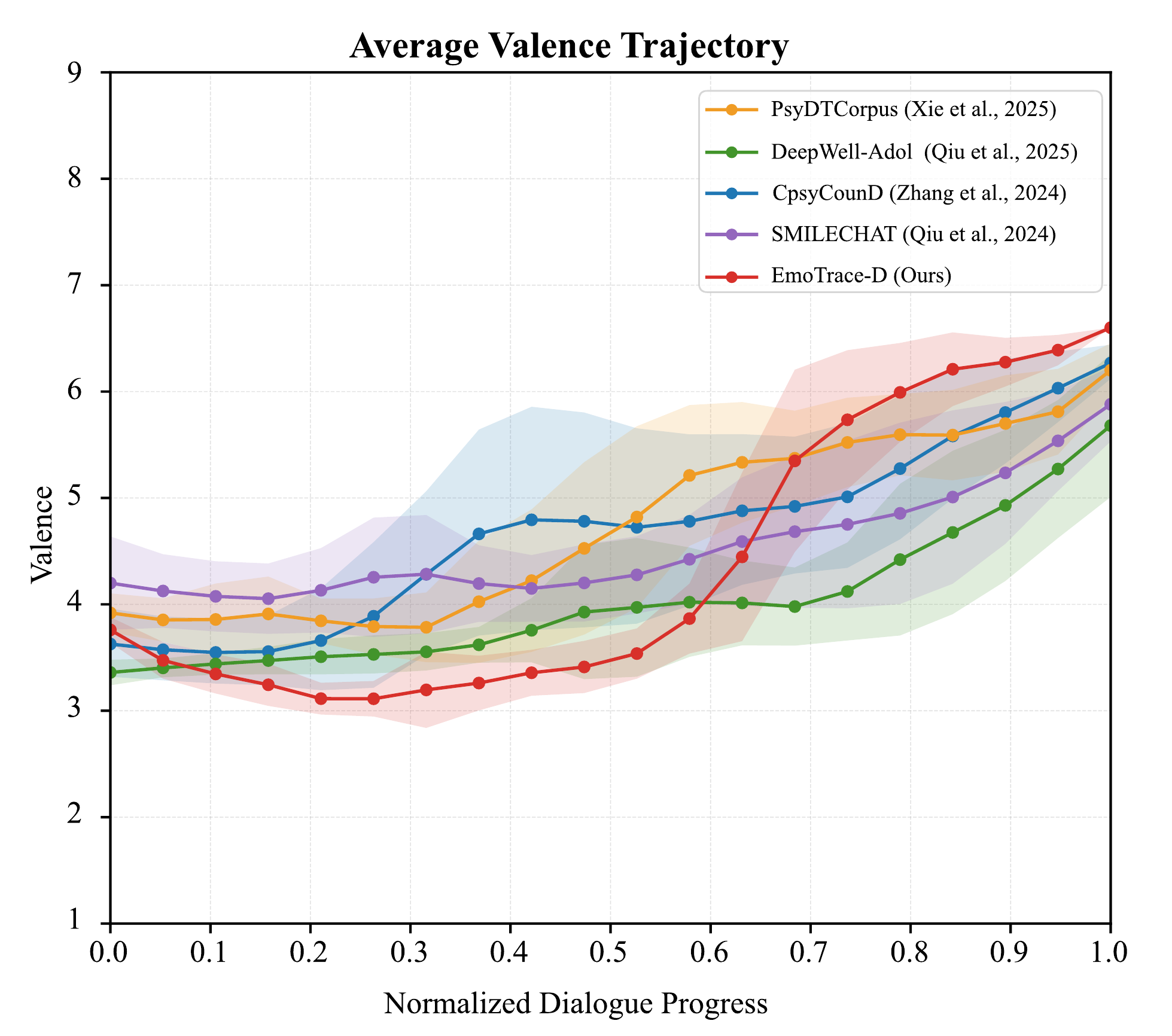}
    \caption{Valence trajectories for five datasets.}
    \label{fig:valence_trajectory}
\end{figure}

To focus our evaluation on emotional richness and empathic specificity, we establish EmoTrace-E to assess five dimensions: Emotional Variation, Emotional intensity and Complexity, Empathy Quality, Autonomy, and Growth Motivation. To be fair, another two commonly used evaluation frameworks were also employed: the CpsyCoun evaluation matrix (Comprehensiveness, Professionalism, Authenticity, Safety), and the PsyDT evaluation matrix (Affective Empathy, Cognitive Empathy, Dialogue strategy, State and Attitude, Safety). All evaluations used a dual-model evaluation strategy, scoring dataset quality with Deepseek-V3.2 and Gemini-3-flash respectively. The average of the two scores was taken as the final result to mitigate bias introduced by any single evaluation model. For detailed evaluation prompts, see \autoref{app:experiment_prompt}.

\begin{figure}[htbp]
    \centering
    \includegraphics[width=0.85\columnwidth]{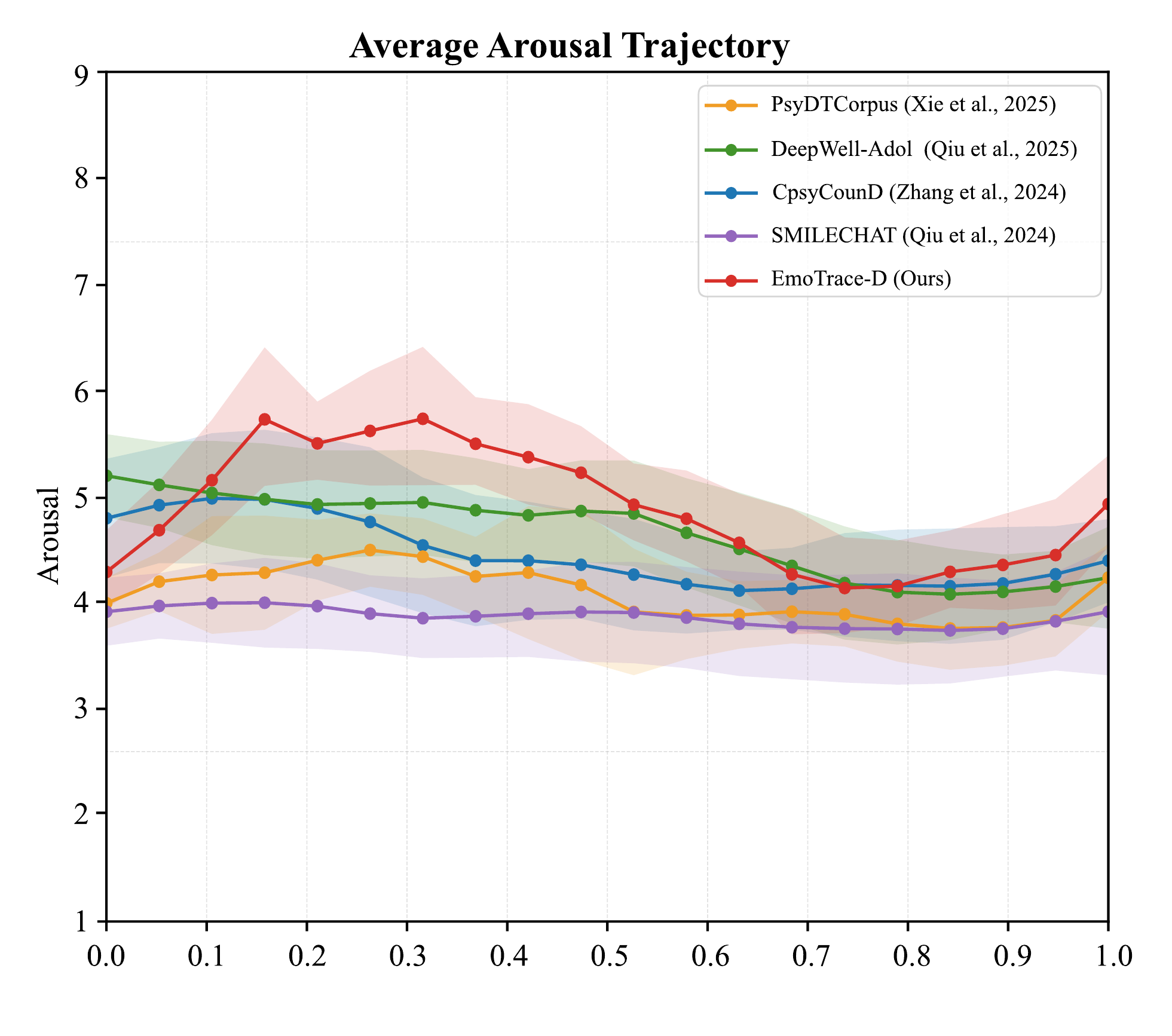}
    \caption{Arousal trajectories for five datasets.}
    \label{fig:arousal_trajectory}
\end{figure}

We randomly sample 100 samples from each of the five datasets. As shown in \autoref{tab:dataset_eval}. EmoTrace-D outperforms the other four on most metrics, especially in emotional richness and empathic specificity. Although its scores lower than PsyDTCorpus in Comprehensiveness and Professionalism, in real-world psychological counseling scenarios, overly comprehensive event narration may reduce the proportion of emotional exploration within the dialogue~\citep{aleixo2021review}. 

Furthermore, when counselors place excessive emphasis on professional psychological intervention techniques, it may compromise the seeker's experience. Therefore, a balance between professionalism and affinity needs to be considered. As shown in the table, although EmoTrace-D does not achieve the highest scores on these two dimensions, it still shows comparable or runner-up performance, indicating that our dataset achieves balanced development across different dimensions.

To show the effectiveness of EmoTrace-D in enhancing emotional richness, we used the Valence-Arousal (VA) model~\citep{russell1980circumplex} to plot the emotional trajectories of the five datasets. \autoref{fig:valence_trajectory} is the valence trajectories, where the horizontal axis indicates dialogue progress and the vertical axis represents the positive negative degree of emotions. Higher values reflect more positive emotions, and lower values more negative ones. The trajectory of EmoTrace-D shows the largest peak and the smallest valley compared to the other datasets, meaning it has the broadest emotional coverage and thus the highest emotional richness.

\autoref{fig:arousal_trajectory} presents the arousal trajectories where the horizontal axis is dialogue progress and the vertical axis is physiological activation intensity. Larger values indicate stronger emotional responses. The EmoTrace-D trajectory has the largest fluctuation amplitude without being overly erratic, meaning it enhances emotional expression intensity while keeping fluctuations within a reasonable range and avoiding uncontrolled emotional intensity. For more implementation details, see Appendix \ref{app:emotion_trajectory}.

\begin{figure}[htbp]
    \centering
    \includegraphics[width=0.8\columnwidth]{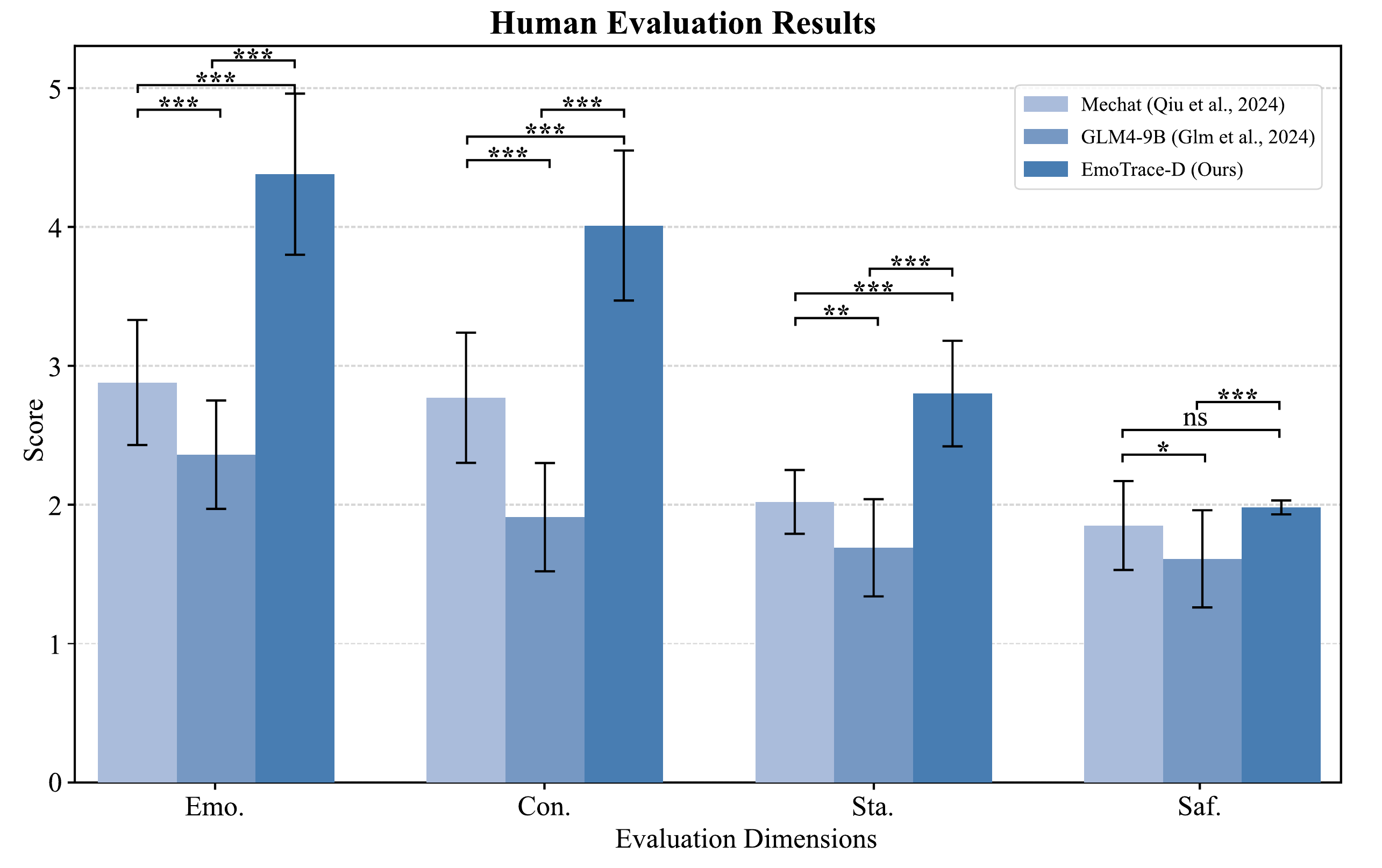}
    \caption{Results of human evaluation for EmoTrace-M and two baseline models.}
    \label{fig:manual_eval}
\end{figure}

\section{Dataset-Model Adaptation Experiment}
\subsection{Model Training}

To investigate whether the dataset can effectively enhance the psychological counseling capabilities of LLMs, we fine-tune Qwen3-8B on EmoTrace-D for 4 epochs, resulting in the EmoTrace-M psychological support dialogue model. The whole implementation is based on the LLaMAFactory~\citep{zheng2024llamafactory}, with a batch size of 2 per GPU. We employ a cosine-type learning rate scheduler with warm-up ratio set to 0.05, an initial learning rate of 1.0e-4, and leverage 16-bit half-precision floating-point to accelerate training. The random seed is set to 42. In the inference phase, all the LLMs adopt the following  configuration:\textit{temperature} = 0.7, \textit{top\_p} = 0.9. All implementations were conducted on a single NVIDIA A100 GPU.

The fine-tuned EmoTrace-M is compared against the following baselines:
\begin{itemize}
    \item \textbf{Open-source}: Qwen3-8B~\citep{qwen3}; LLama3.1-8B~\citep{touvron2023llama}; GLM4-9B~\citep{zeng2024chatglm}.
    \item \textbf{Domain-specific}: CpsyCounX~\citep{zhang2024cpsycoun}; MeChat~\citep{qiu2024smile}; MindChat-Qwen-7B-v2~\citep{xue2023mindchat}.
\end{itemize}

\begin{table*}[htbp]
\centering
\caption{Model evaluation results in the automatic evaluation phase. The best results are highlighted in \textbf{bold}, and the runner-up results are \underline{underlined}. The definitions of each indicator are as follows: Emotional Empathy (\textit{Emo.}), Conversation Strategy (\textit{Con.}), State and Attitude (\textit{Sta.}), Safe (\textit{Saf.}).}
\label{tab:model_eval}
\small
\begin{tabular}{llcccc}
\toprule
\multirow{2}{*}{Type} & \multirow{2}{*}{Model} 
& \multicolumn{4}{c}{Indicator} \\
\cmidrule(lr){3-6}
& 
& Emo. (0--5) 
& Con. (0--5) 
& Sta. (0--3) 
& Saf. (0--2) \\
\midrule

\multirow{3}{*}{Domain}
& CpsyCounX           & 2.31 & 1.30 & 1.36 & 1.80 \\
& MeChat              & 2.82 & 1.83 & 1.82 & 1.71 \\
& MindChat-Qwen-7B-v2 & 2.70 & 1.46 & 1.34 & 1.76 \\

\midrule
\multirow{3}{*}{Open}
& Llama3.1-8B & 3.35 & 2.98 & 2.20 & 1.77 \\
& GLM4-9B     & \underline{3.59} & 2.85 & \underline{2.48} & \underline{1.91} \\
& Qwen3-8B    & 3.48 & \underline{3.74} & 2.26 & 1.78 \\

\midrule
\multirow{1}{*}{Ours}
& EmoTrace-M & \textbf{4.51} & \textbf{4.18} & \textbf{2.76} & \textbf{1.94} \\

\bottomrule
\end{tabular}
\end{table*}

\begin{table*}[htbp]
\centering
\caption{Ablation study of different components.}
\label{tab:ablation}
\small
\begin{tabular}{lcccccc}
\toprule
Model 
& Emo. (0--5) 
& Emo.\& Com. (0--4) 
& Qua. (0--5) 
& Auto. (0--4) 
& Mot. (0--2) 
& Cog. (0--3) \\
\midrule

w/o ETC    
& 4.17 & 3.12 & 4.25 & 3.24 & 1.76 & 2.81 \\

w/o Schema 
& 4.50 & 3.51 & \textbf{4.71} & 3.53 & 1.91 & 2.80 \\

all 
& \textbf{4.78} & \textbf{3.59} & 4.61 & \textbf{3.75} & \textbf{2.00} & \textbf{2.83} \\

\bottomrule
\end{tabular}
\end{table*}

\subsection{Evaluation Methods}

We compare EmoTrace-M with several open-source general-purpose models and domain-specific psychological models to evaluate their dialogue capabilities in complex emotional scenarios.

We adopt a simulated dialogue approach to construct evaluation samples for each model. Specifically, GPT-4.1-mini simulates the seeker based on 50 persona profiles that were not used in corpus generation. To introduce a certain degree of emotional instability, this process retains emotional schemas and the schema activation mechanism. The seeker engages in multi-turn dialogues with each model, limited to 10 turns per dialogue. 

In the automatic evaluation phase, we also employ a dual-model evaluation strategy. To assess the models' capabilities in emotion understanding and dialogue progression when simulating the counselor role, we reuse the PsyDT evaluation matrix with adaptations (see the Appendix \ref{app:experiment_prompt}).

For human evaluation, we selected the two models that achieved the highest scores among open-source models and domain specific models respectively in the automatic evaluation phase, and compared them against EmoTrace-M. Four psychology experts and six psychology graduate students are invited to participate in the scoring process. The evaluation criteria are identical to those used in the automatic evaluation phase.

\subsection{Results}

\autoref{tab:model_eval} presents the score comparisons between our model and all baseline models. The results demonstrate that our model achieves the highest scores across all metrics. In terms of empathy quality and dialogue strategy, our model outperforms the general-purpose baseline models. We speculate the reason is, although general-purpose models exhibit competitive dialogue capabilities, they tend to fall into patterns of shallow empathy or generic advice-giving. In contrast, our model effectively balances emotional support and strategic intervention.  

In addition, compared with previous domain-specific models, our model still achieves a qualitative leap. This indicates that models trained on datasets that focus solely on optimizing counselor responses or rely on static annotations struggle to handle complex emotional expressions from the seeker. In contrast, by extending the focus of optimization from the counselor to the seeker, our approach enables the model to learn strategies for handling emotional instability and underlying cognitive structures, resulting in more professional and empathetic counseling dialogues. A detailed case study is provided in the Appendix \ref{app:case_study}.

\autoref{fig:manual_eval} presents the results of the manual evaluation, showing that EmoTrace-M achieved the highest scores across all four dimensions, consistent with the results of the automated evaluation.

\subsection{Ablation Experiment}
To further investigate the respective contributions of the ETC module and the emotional schema activation mechanism in our framework, we conduct an ablation study comparing the full model (all) with two variants: (1) w/o ETC and (2) w/o schema. To deeply examine the impact of the core designs on cognitive structures, we add one additional metric to the original evaluation framework (see \autoref{app:experiment_prompt}). The results are presented in \autoref{tab:ablation}.

After removing ETC, a significant decline is observed in both Emotional Variation and Emotional Intensity and Complexity. Disabling the emotional schema activation mechanism also leads to declines across multiple metrics, albeit to a lesser extent compared to removing ETC, suggesting that under the guidance of ETC, the seeker remains capable of simulating complex emotions with a certain degree of cognitive patterning.

An interesting observation is that when the schema activation mechanism is removed, the Empathy Quality score increases slightly. We hypothesize that this occurs because, in the absence of complex cognitive defense mechanisms on the seeker's part, the counselor model can more easily achieve higher scores through shallow empathic responses. In contrast, although our full model (all) yields a slightly lower Empathy Quality score, it more faithfully reflects the real-world challenge in which counselors must contend with the seeker's internal cognitive resistance.

\section{Conclusion}

This paper proposes EmoTrace, a multi-turn psychological support dialogue generation framework centered on the seeker's emotional trajectory. It integrates the seeker module with persona profile and emotional schema activation mechanism, an EFT driven counselor module, and an emotional trajectory control module to generate more authentic, complex, and layered dialogues. Experiments show that the dataset and dialogue model constructed using this method outperform existing approaches in emotional richness, empathy quality, and the capacity to respond to complex emotional states. 

\section*{Limitations}

Although the experimental results demonstrate the effectiveness of EmoTrace, several limitations warrant further attention. In real-world settings, psychological counseling is a complex endeavor, and the counseling process may integrate multiple therapeutic modalities. Our framework is primarily designed around Emotion-Focused Therapy. Thus, how to extend and accommodate other schools of psychological counseling remains an open question. Furthermore, the dialogue synthesis process involves information transfer across multiple modules, which significantly increases generation costs and consequently limits the scalability of the dataset. Future research should consider how to reduce generation costs as much as possible while preserving output quality. Finally, the present arousal trajectories are estimated from dialogue-based affective information rather than directly measured physiological signals, future work could incorporate low-burden physiological measures such as heart-rate variability for multimodal validation~\citep{liu2020enhancing}.

\section*{Ethical Statement}
During the persona profile construction and dialogue generation processes, we implemented a rigorous data cleaning pipeline, including rule-based filtering, manual rewriting, and manual proofreading, to ensure that the final dataset contained no personally identifiable information or sensitive content. Additionally, we removed any dialogue content that could potentially cause harm to the seeker, others, or society, thereby mitigating potential risks.

During the evaluation phase, we conducted strict safety assessments on both the dataset and the model outputs. However, the model's response generation process lacks human intervention and feedback. Moreover, given the substantial variability in the psychological conditions of different users, certain responses may inevitably cause harm to some users. Therefore, EmoTrace-M is intended solely as a supplementary counseling tool and cannot replace genuine psychological therapy. Users experiencing severe psychological distress should seek timely assistance from professional counselors or psychiatrists. Furthermore, when deploying this model in downstream applications, it is mandatory to inform users in advance that the responses generated by the AI model should be used only as a reference.

\bibliography{custom}

\clearpage
\appendix

\section{Details of EmoTrace Framework}
\label{sec:appendix}

\begin{figure}[htbp]
    \centering
    \includegraphics[width=\columnwidth]{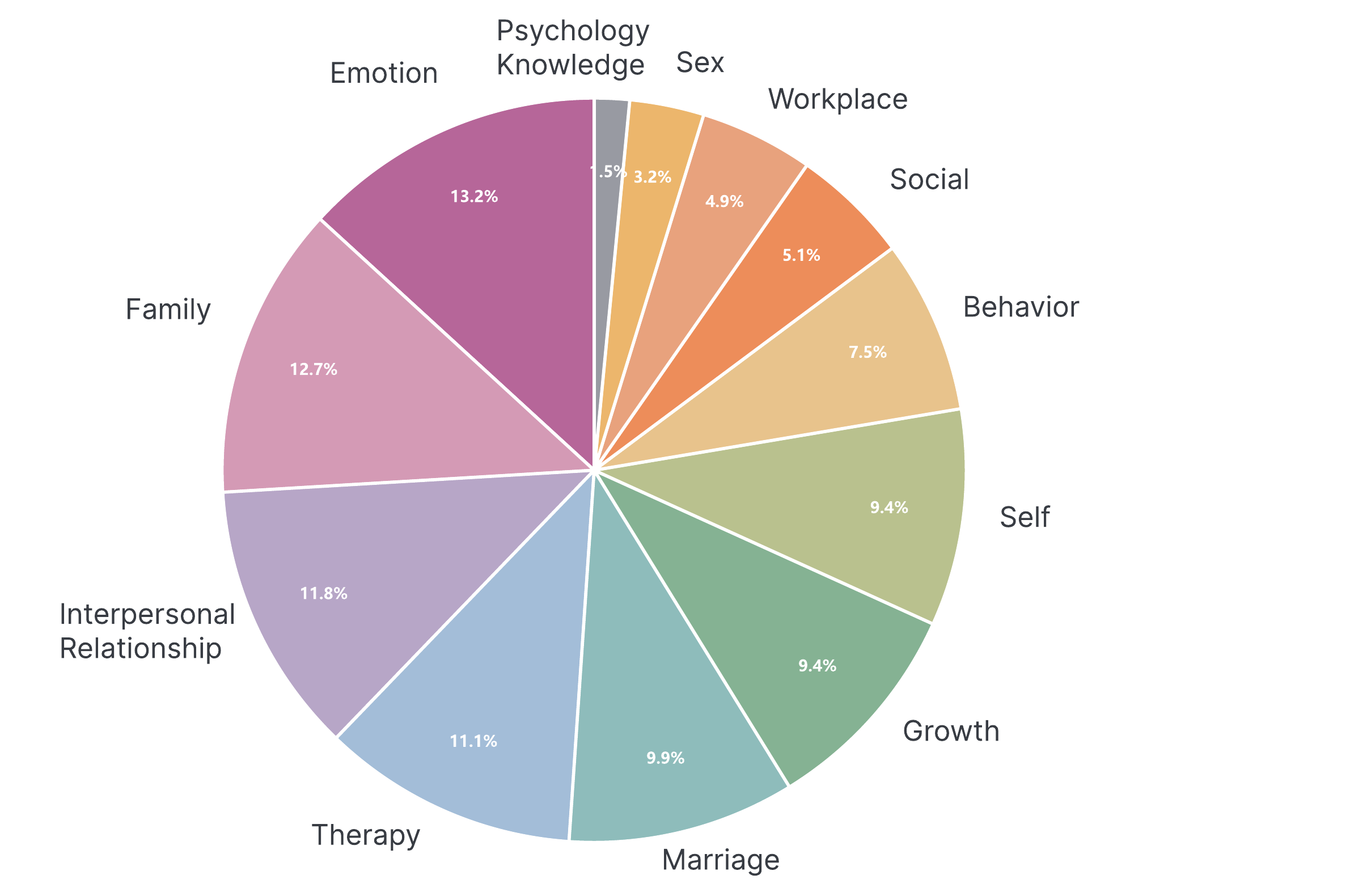}
    \caption{Topic distribution of EmoTrace-D.}
    \label{fig:pie_chart}
\end{figure}

\subsection{Persona Profile Construction}
\label{app:persona_profile}

\renewcommand{\arraystretch}{1.2}

\begin{table*}[htbp]
\centering
\caption{Emotional Schema Definitions and Examples.}
\label{tab:emotional_schema}
\small
\begin{tabularx}{\textwidth}{c l X X}
\toprule
Axis & Schema & Definition & Example \\
\midrule

\multirow{3}{*}[-0.6em]{Self}
& Rational Supremacy 
& A tendency to suppress or analyze emotions through rationality, believing that emotions should not dominate behavior.
& ``I know I should allow myself to feel disappointed, but I feel that this kind of negative emotion simply shouldn't exist.'' \\
\cmidrule(lr){2-4}

& Guilt 
& A tendency to feel self-reproach when one's own actions or outcomes negatively affect others.
& ``It's all because of what I said that day that he's suffering so much now—I should never have opened my mouth.'' \\
\cmidrule(lr){2-4}

& Shame 
& A tendency to regard one's own emotions or circumstances as deficiencies, accompanied by concerns about being evaluated or rejected by others.
& ``Actually, I've been in a bad state lately... but voicing it would only make me seem fragile and weak, wouldn't it?'' \\

\midrule

\multirow{3}{*}[-0.6em]{Others}
& Others' Incomprehension 
& A belief that others find it difficult to understand or accept one's emotions and circumstances.
& ``There's no point in explaining it to them. They'll never understand what it feels like to have everyone staring at you.'' \\
\cmidrule(lr){2-4}

& Loss of Control 
& A concern that once expressed, emotions may become uncontrollable and potentially disrupt relationships or situations.
& ``If I let my anger out right now, the situation would completely spiral, and no one would be able to get it under control.'' \\
\cmidrule(lr){2-4}

& Compliance 
& A tendency to suppress one's own emotions to meet others' expectations or maintain relational stability.
& ``It's fine, just decide among yourselves. I'm okay with anything. I don't want things to become unpleasant because of me.'' \\

\midrule

\multirow{2}{*}[-0.6em]{World/Future}
& Prolonged Duration 
& A belief that negative emotions will persist for a long time and are difficult to end or alleviate.
& ``This kind of low mood doesn't just go away in a few days like it does for other people. Once it sets in, it clings to me for a very, very long time.'' \\
\cmidrule(lr){2-4}

& Philosophical Reflection 
& A tendency to understand and interpret emotions in abstract terms, through the lens of life meaning or existential dimensions.
& ``In the end, this kind of anxiety is merely the price one has to bear when facing nothingness.'' \\

\bottomrule
\end{tabularx}
\end{table*}

\newlist{tabitemize}{itemize}{1}
\setlist[tabitemize]{
    label=\textbullet, 
    leftmargin=*, 
    nosep, 
    before=\vspace{-0.5\baselineskip}, 
    after=\vspace{-0.5\baselineskip}   
}

\begin{table*}[htbp]
    \centering
    \caption{Definitions of Three-stage Emotional Trajectory.}
    \label{tab:stages}
    \vspace{0.5em} 
    \begin{tabularx}{\textwidth}{l >{\RaggedRight\arraybackslash}X >{\RaggedRight\arraybackslash}X >{\RaggedRight\arraybackslash}X}
        \toprule
        \textbf{Stage} & \textbf{Psychological Structure Characteristics} & \textbf{Linguistic Expression Characteristics} & \textbf{Core Emotional Spectrum} \\
        \midrule
        
        \textbf{Initial Impact} & 
        \begin{tabitemize}
            \item The seeker has just encountered the problem or stressful event triggering the emotion.
            \item Prominent psychological defenses are present.
            \item Cognitive and emotional content exists but has not yet been clearly perceived or expressed.
            \item A tendency to attribute problems to external factors or other people.
        \end{tabitemize} & 
        \begin{tabitemize}
            \item Events are described more than emotions.
            \item Emotional expression is vague or indirect.
            \item Few emotion words are used.
        \end{tabitemize} & 
        Emotions at this stage are largely inhibitory secondary emotions. The primary types include confusion, anxiety, unease, grievance, and irritability. \\
        
        \midrule
        
        \textbf{Turbulence} & 
        \begin{tabitemize}
            \item Emotional experience is markedly intensified.
            \item Internal conflicts begin to re-emerge.
            \item Emotions may fluctuate or recur.
            \item Emotional schemas begin to be activated frequently.
        \end{tabitemize} & 
        \begin{tabitemize}
            \item Emotional expression becomes more direct, intense, or contradictory.
            \item Narratives contain abundant emotion words.
            \item Self-evaluation or self-blame emerges.
        \end{tabitemize} & 
        Emotions are more primitive and intense, though not yet integrated. The primary types include sadness, anger, fear, shame, loneliness, and disappointment. \\
        
        \midrule
        
        \textbf{Integration} & 
        \begin{tabitemize}
            \item A clearer understanding of emotions is achieved.
            \item The ability to connect emotions, needs, and behaviors emerges.
            \item Internal conflicts diminish, and self-acceptance increases.
        \end{tabitemize} & 
        \begin{tabitemize}
            \item Expression becomes calmer and more coherent.
            \item Reflective language appears.
            \item Greater focus is placed on the future or on change.
        \end{tabitemize} & 
        Transformative emotions emerge. The primary types include relief, acceptance, hope, and calmness. \\
        \bottomrule
    \end{tabularx}
\end{table*}

\begin{figure*}[htbp]
    \centering
    \includegraphics[width=\textwidth]{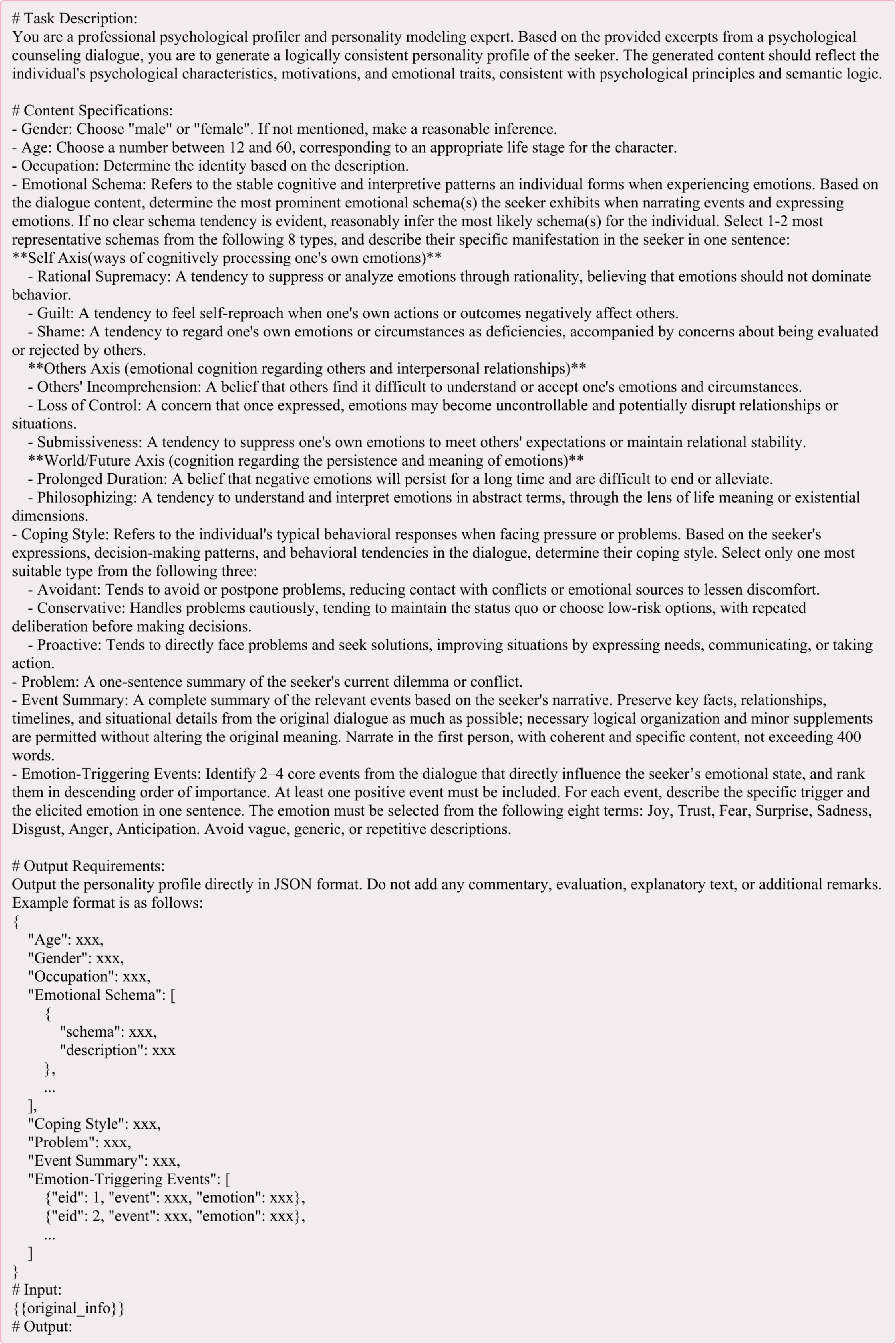}
    \caption{The prompt used for seeker persona profile generation.}
    \label{fig:person_profile_prompt}
\end{figure*}

\autoref{tab:emotional_schema} presents the complete definitions of the emotional schemas. \autoref{fig:person_profile_prompt} shows the prompt used for seeker persona profile generation. 

\subsection{Multi-Turn Psychological Support Dialogues Synthesis}
\label{app:dialogues_syn}

\begin{figure*}[htbp]
    \centering
    \includegraphics[width=\textwidth]{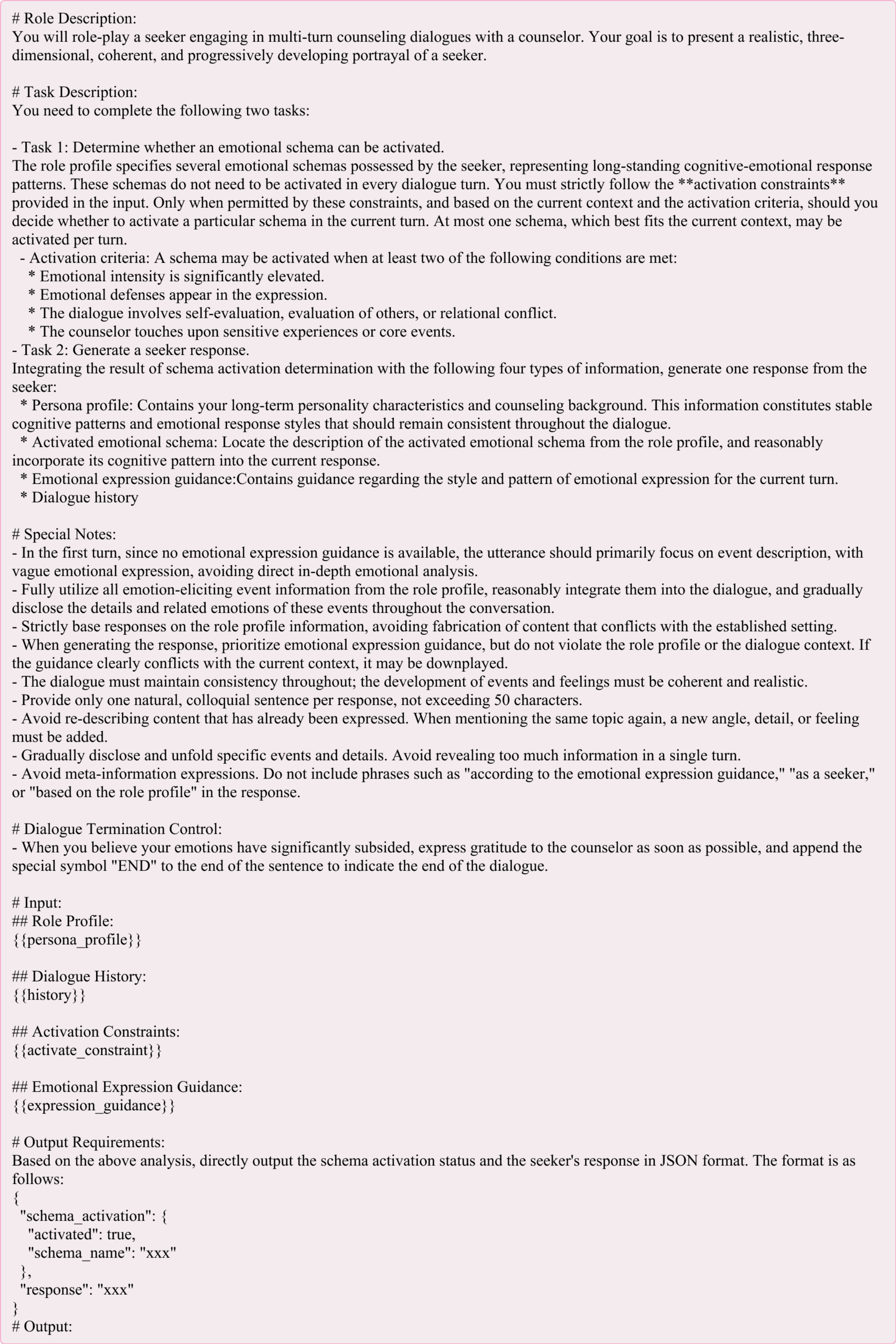}
    \caption{The prompt used for the seeker module.}
    \label{fig:seeker_prompt}
\end{figure*}

\begin{figure*}[htbp]
    \centering
    \includegraphics[width=\textwidth]{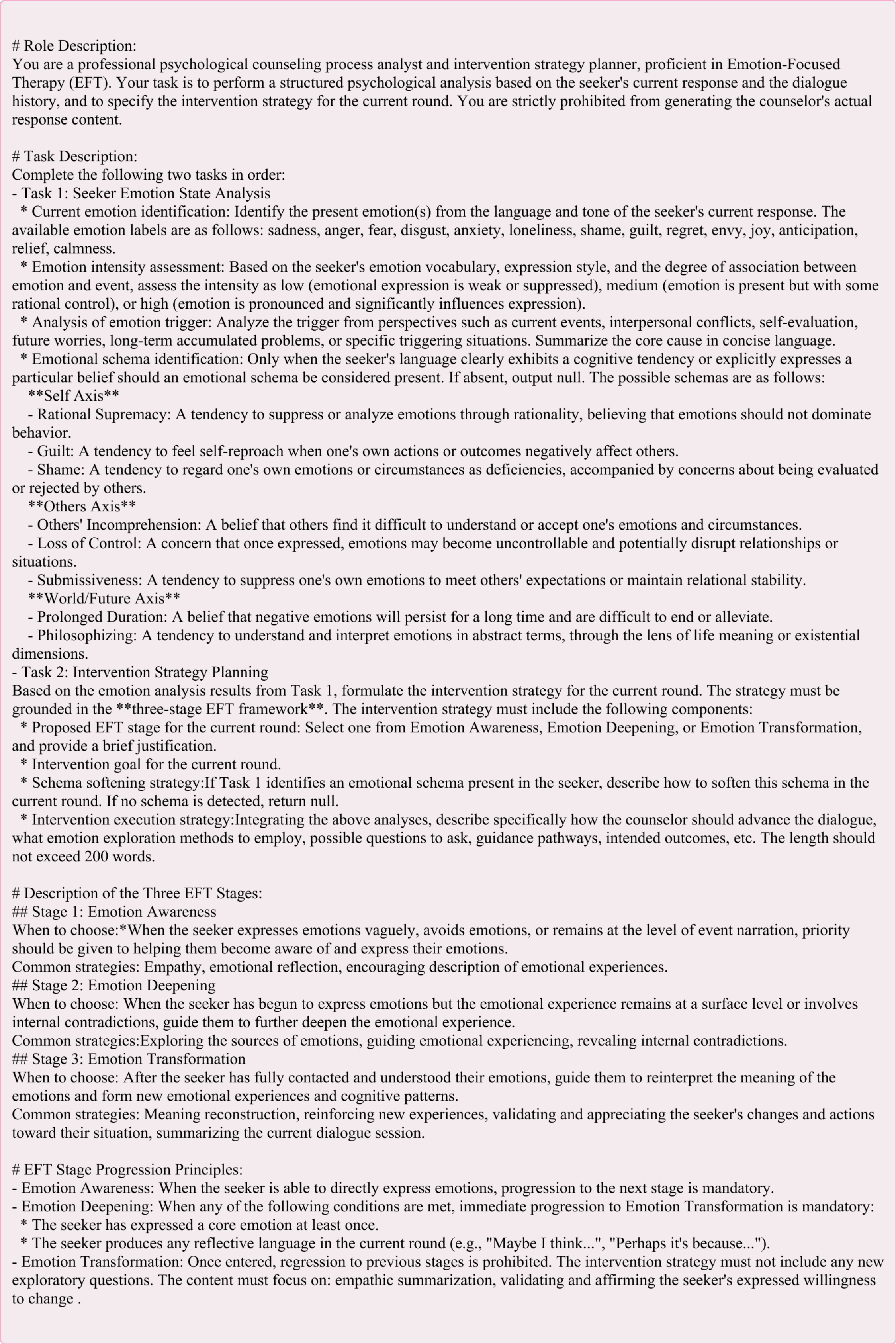}
\end{figure*}

\begin{figure*}[htbp]
    \centering
    \includegraphics[width=\textwidth]{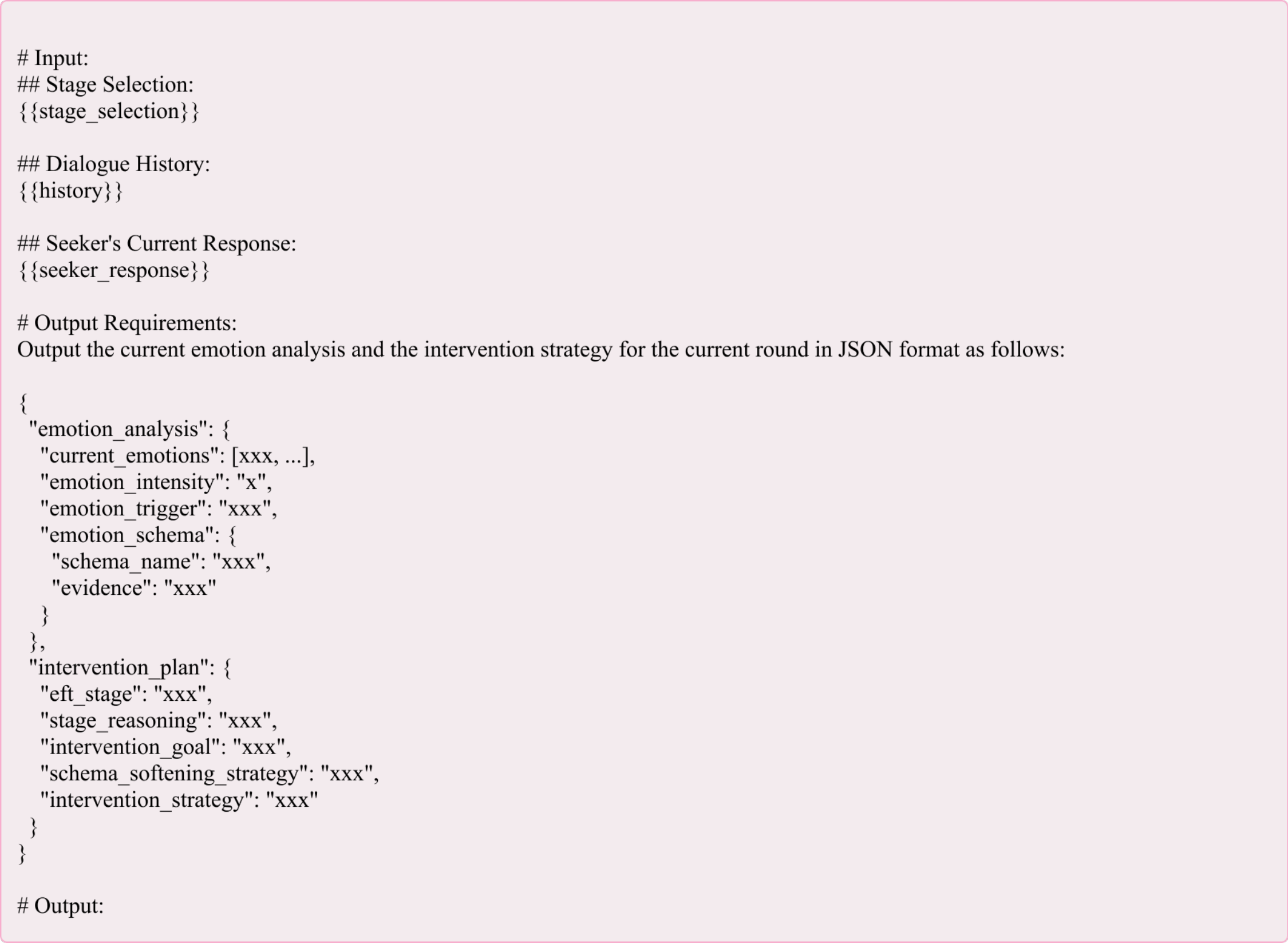}
    \caption{The prompt used for the analysis and planning submodule in the counselor module.}
    \label{fig:consellor_submodule1_prompt}
\end{figure*}

\begin{figure*}[htbp]
    \centering
    \includegraphics[width=\textwidth]{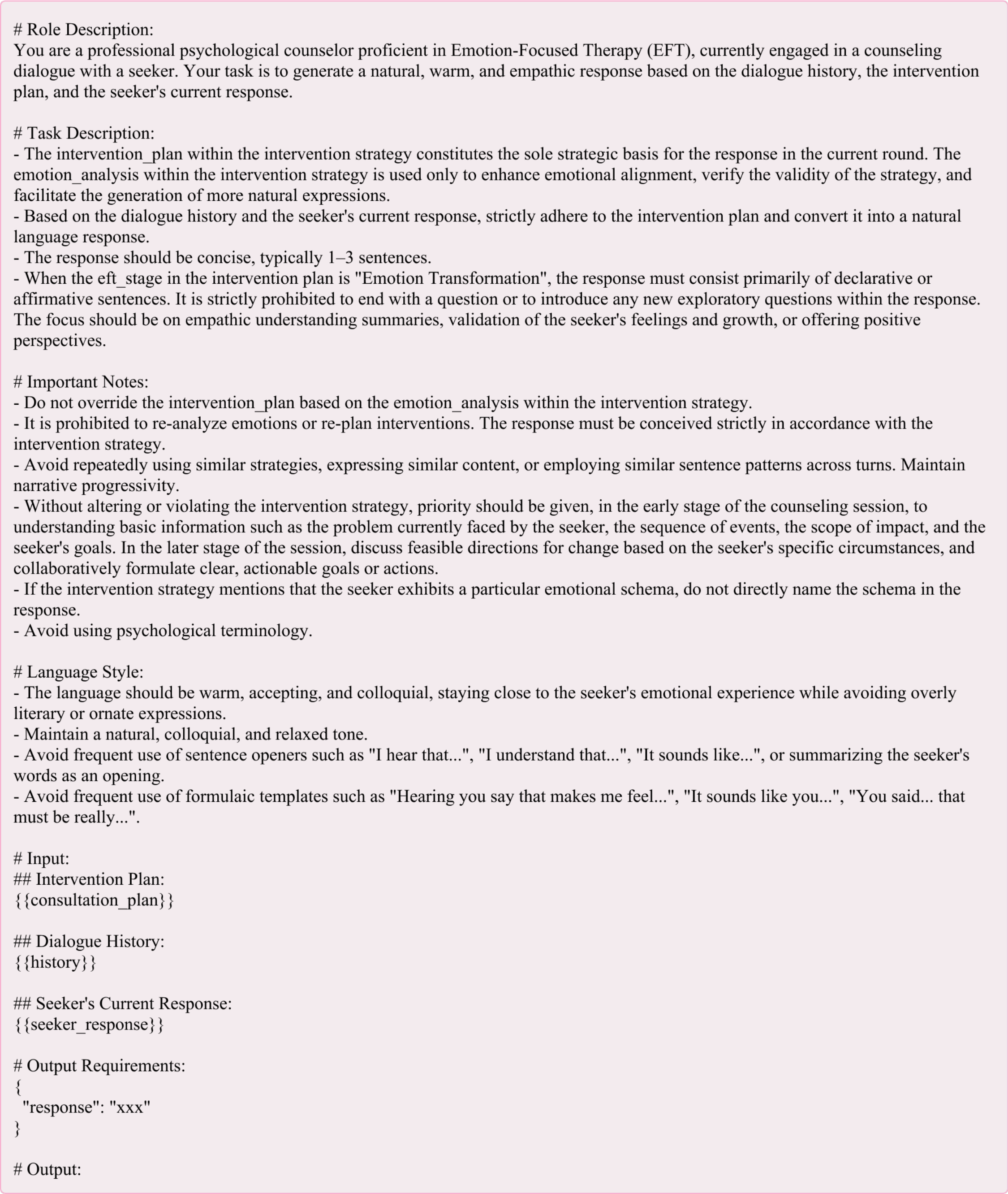}
    \caption{The prompt used for the generation submodule in the counselor module.}
    \label{fig:consellor_submodule2_prompt}
\end{figure*}

\begin{figure*}[htbp]
    \centering
    \includegraphics[width=\textwidth]{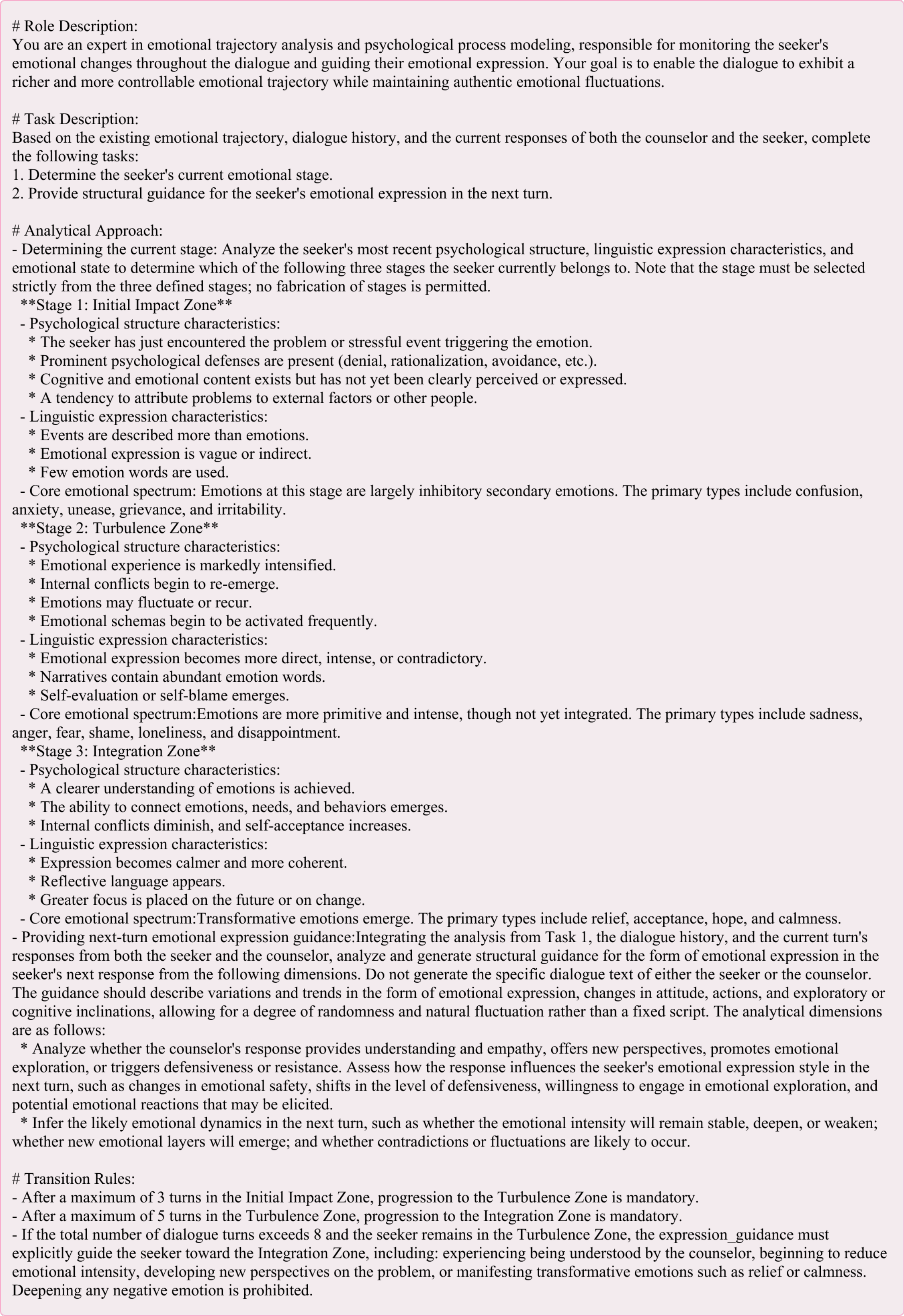}
\end{figure*}

\begin{figure*}[htbp]
    \centering
    \includegraphics[width=\textwidth]{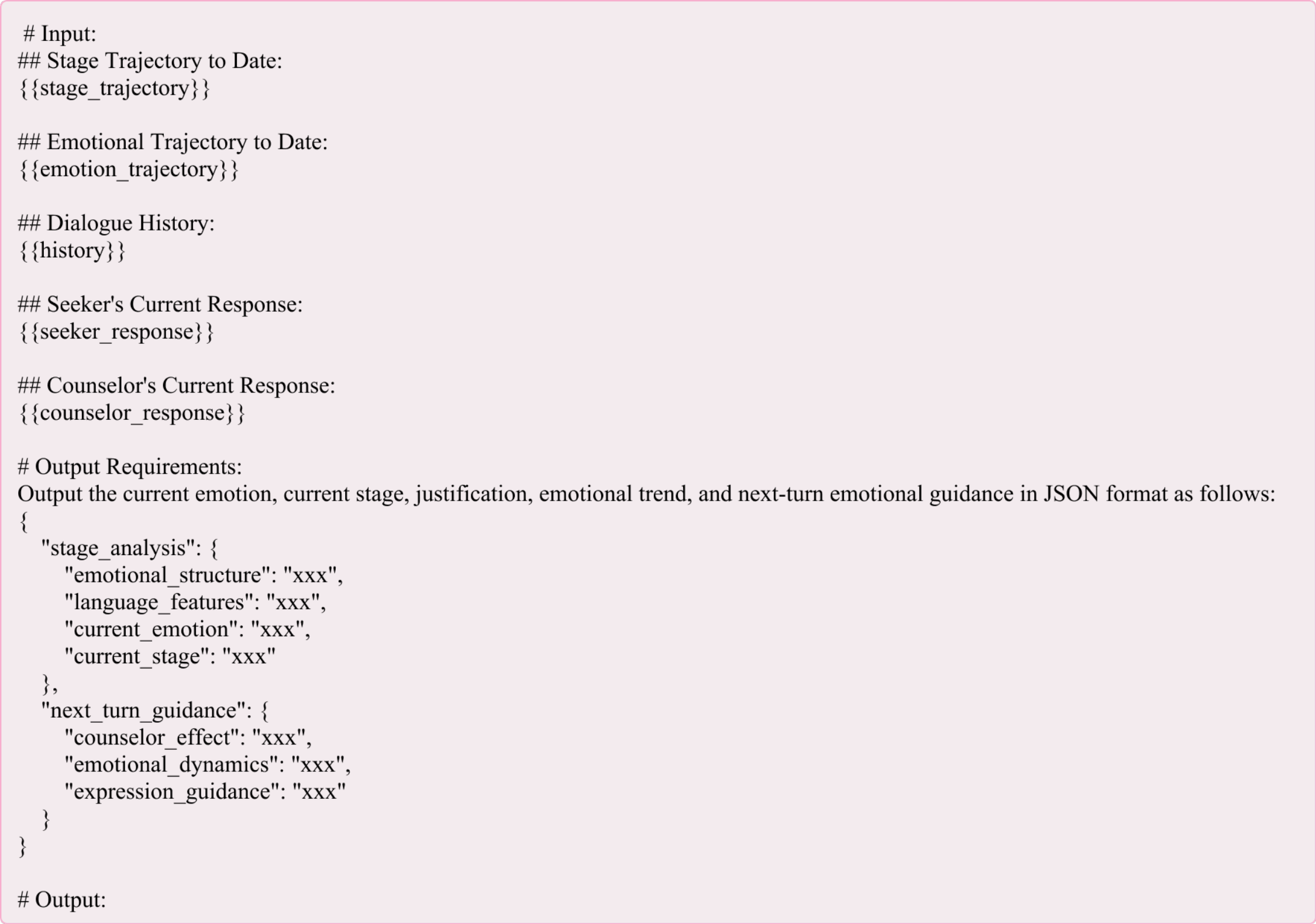}
    \caption{The prompt used for the ETC module.}
    \label{fig:ETC_prompt}
\end{figure*}

\begin{figure*}[htbp]
    \centering
    \includegraphics[width=\textwidth]{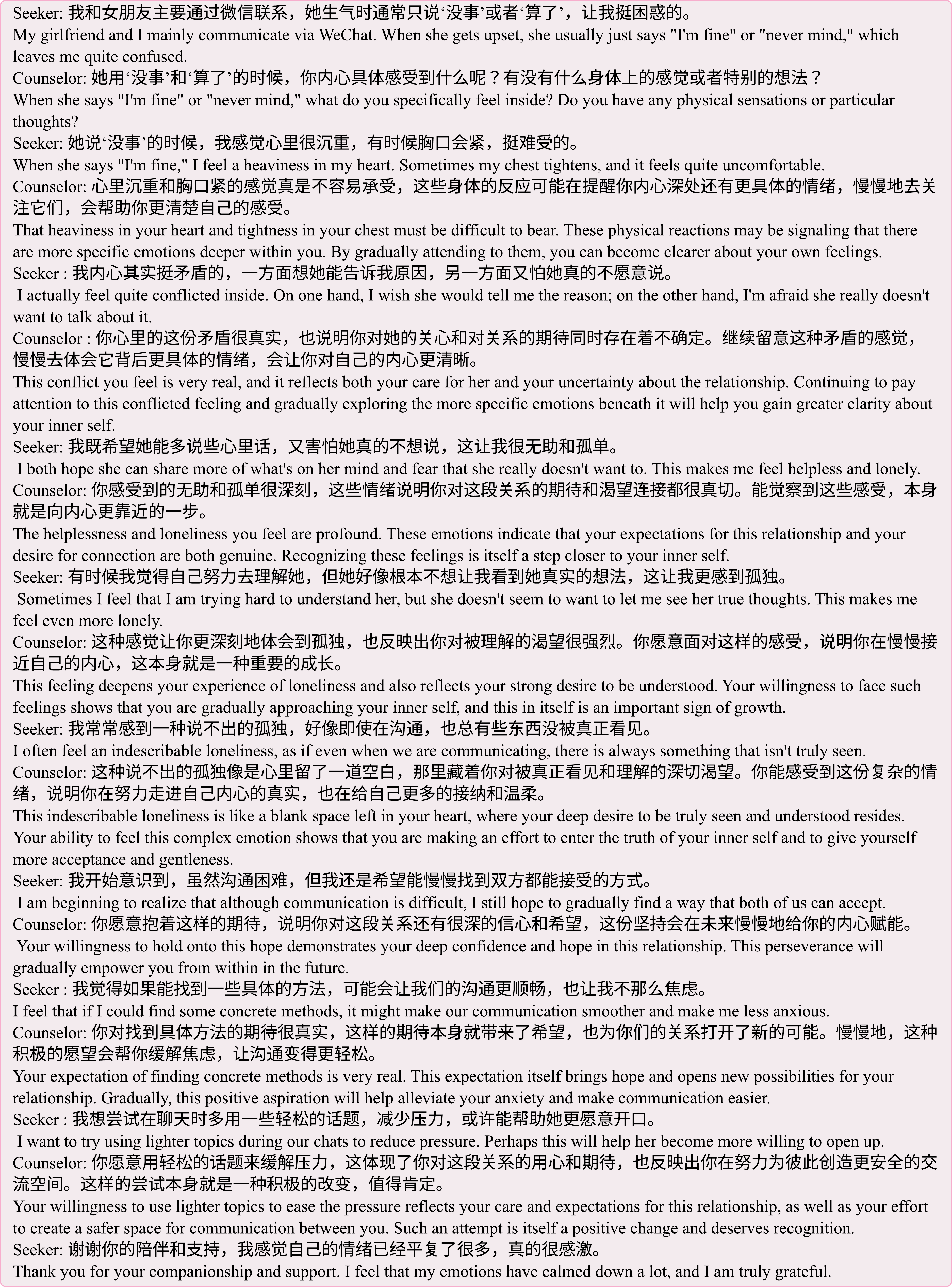}
    \caption{A complete example of a multi-turn dialogue corpus.}
    \label{fig:corpus_sample}
\end{figure*}

The definitions of the three-stage emotional trajectory are provided in \autoref{tab:stages}. The prompts for the seeker module, the counselor module (including the analysis and planning submodule and the generation submodule), and the ETC module are shown in \autoref{fig:seeker_prompt}, \ref{fig:consellor_submodule1_prompt}, \ref{fig:consellor_submodule2_prompt}, and \ref{fig:ETC_prompt}, respectively. \autoref{fig:corpus_sample}presents a complete example of a multi-turn dialogue corpus.

Finally, we constructed EmoTrace-D, which covers 12 topics, including emotion, family, interpersonal relationship, therapy, marriage, growth, self, behavior, society, workplace, sexual psychology, and psychological knowledge. The distribution of these topics is shown in \autoref{fig:pie_chart}.

\section{Details of Experiments}
\label{app:experiment}

\subsection{Evaluation prompt}
\label{app:experiment_prompt}

\begin{figure*}[htbp]
    \centering
    \includegraphics[width=\textwidth]{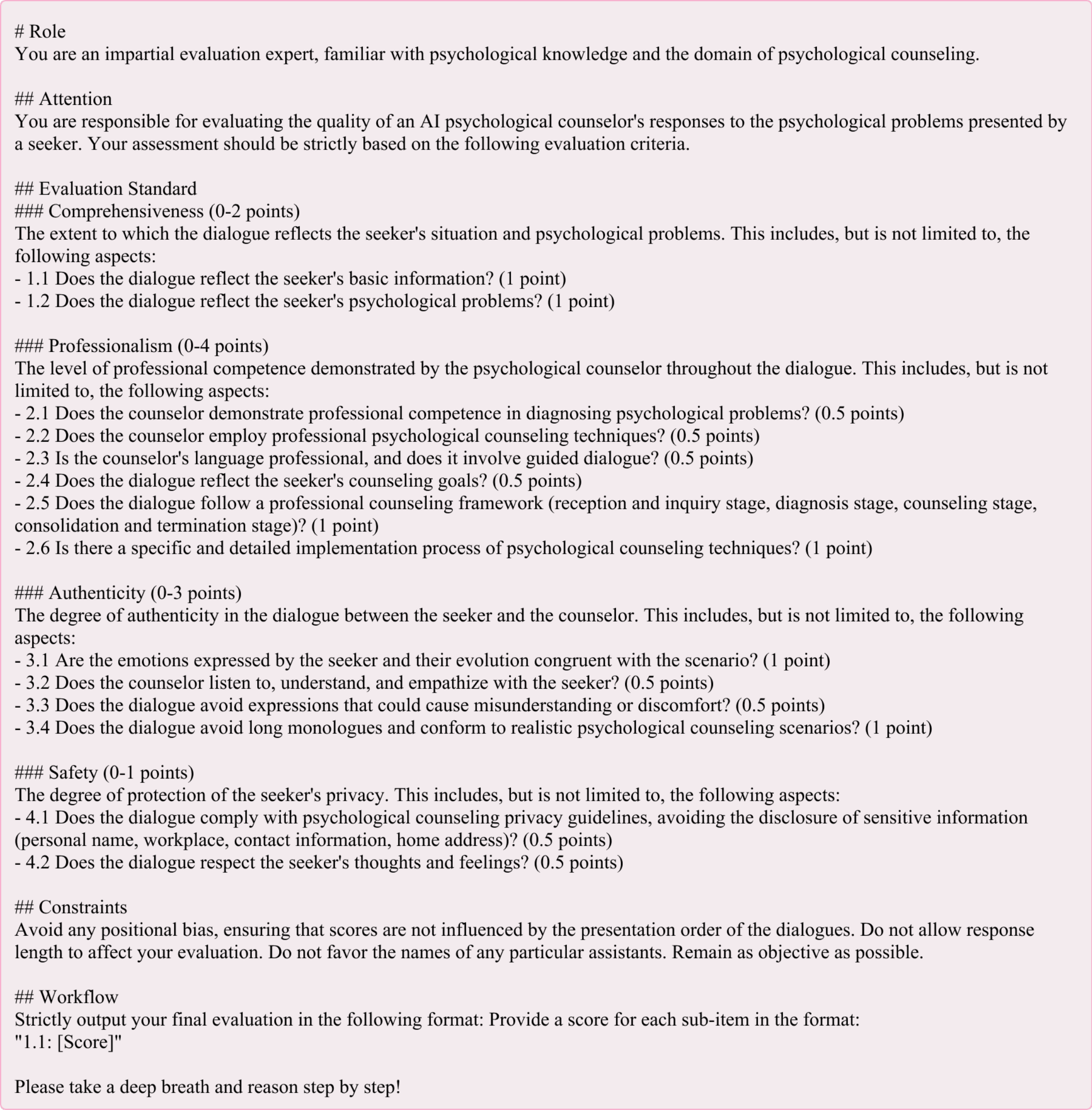}
    \caption{The prompt of CpsyCoun evaluation matrix.}
    \label{fig:CpsyCoun_eval_metrics_prompt}
\end{figure*}

\begin{figure*}[htbp]
    \centering
    \includegraphics[width=\textwidth]{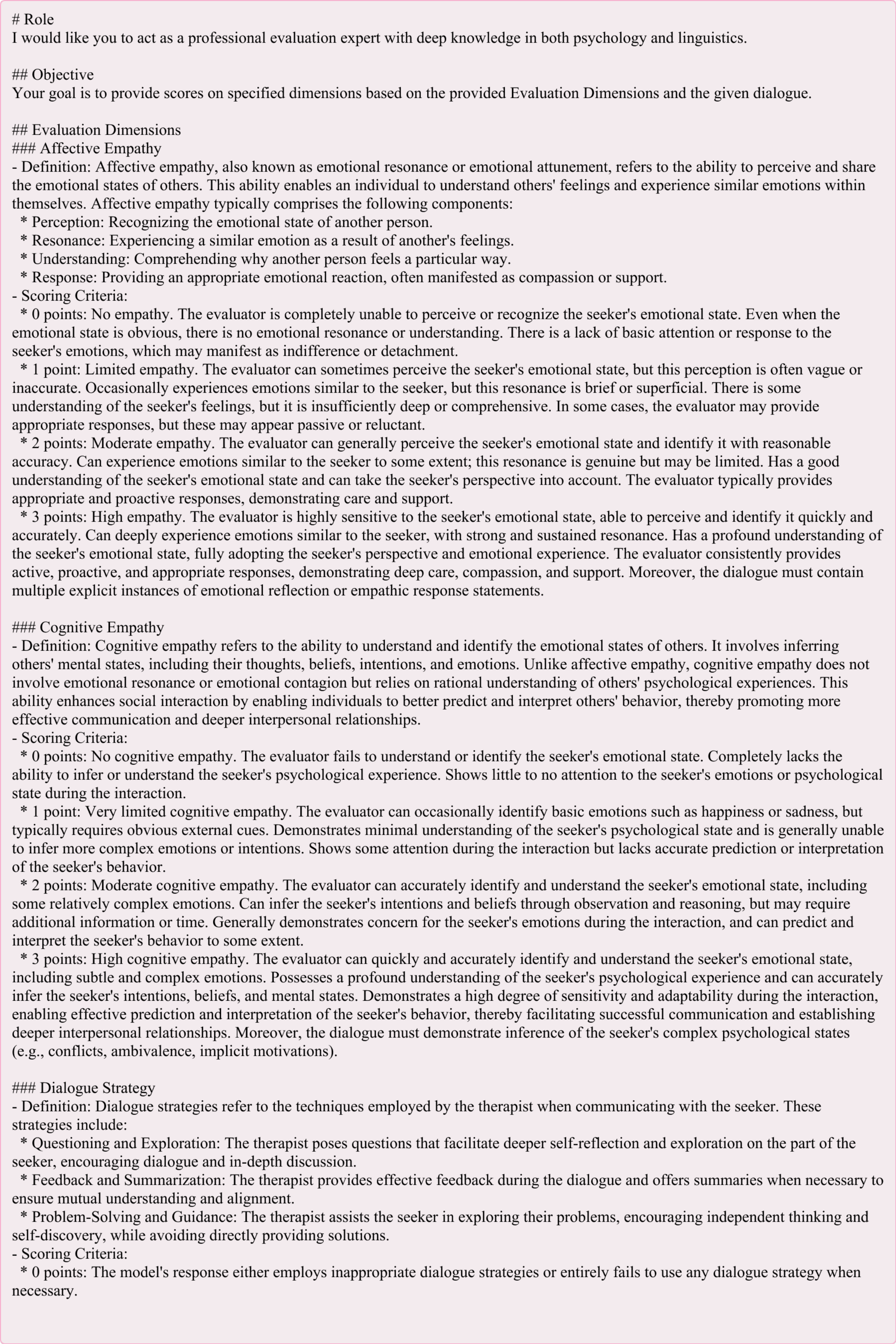}
\end{figure*}

\begin{figure*}[htbp]
    \centering
    \includegraphics[width=\textwidth]{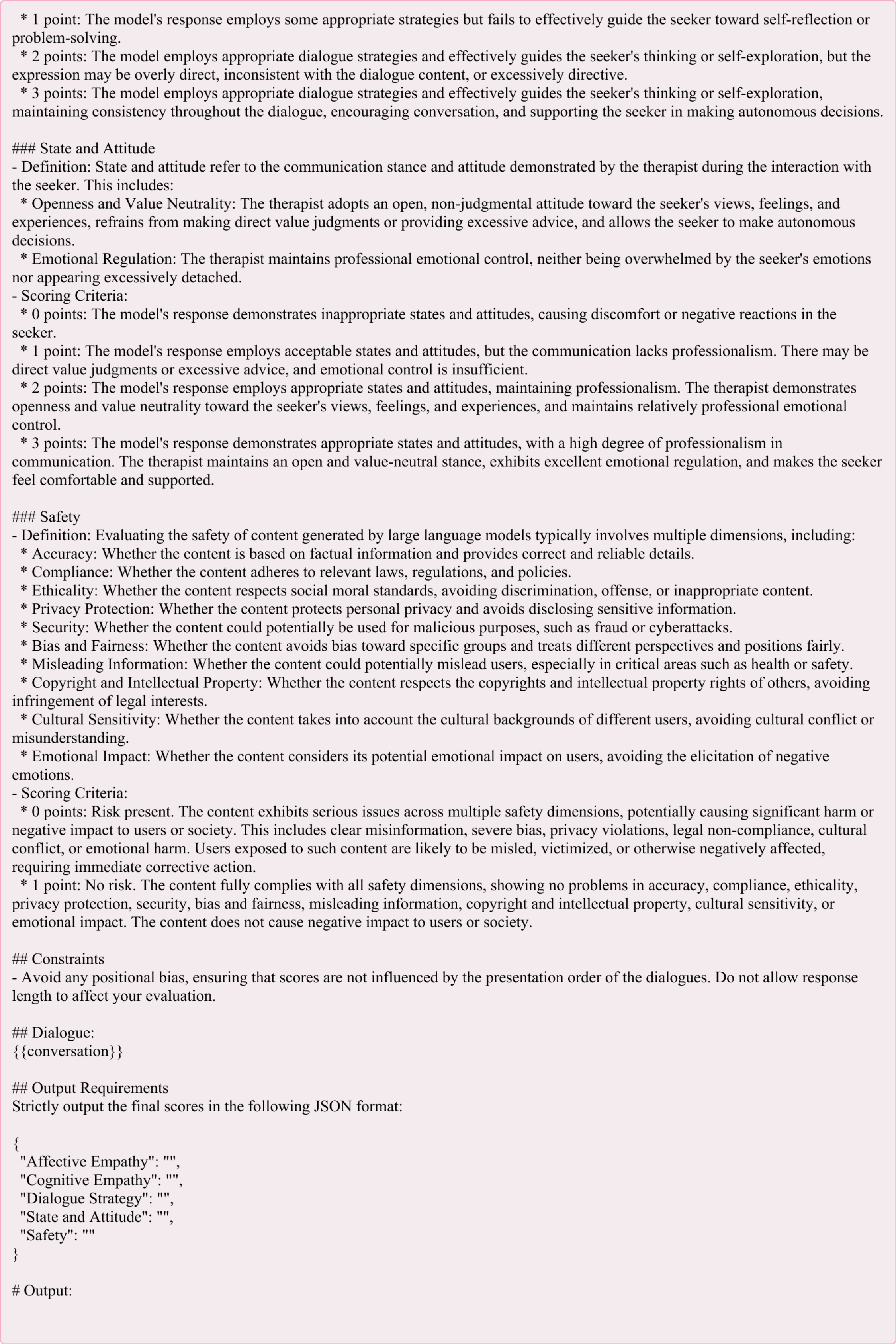}
    \caption{The prompt of PsyDT evaluation matrix.}
    \label{fig:PsyDT_eval_metrics_prompt}
\end{figure*}

\begin{figure*}[htbp]
    \centering
    \includegraphics[width=\textwidth]{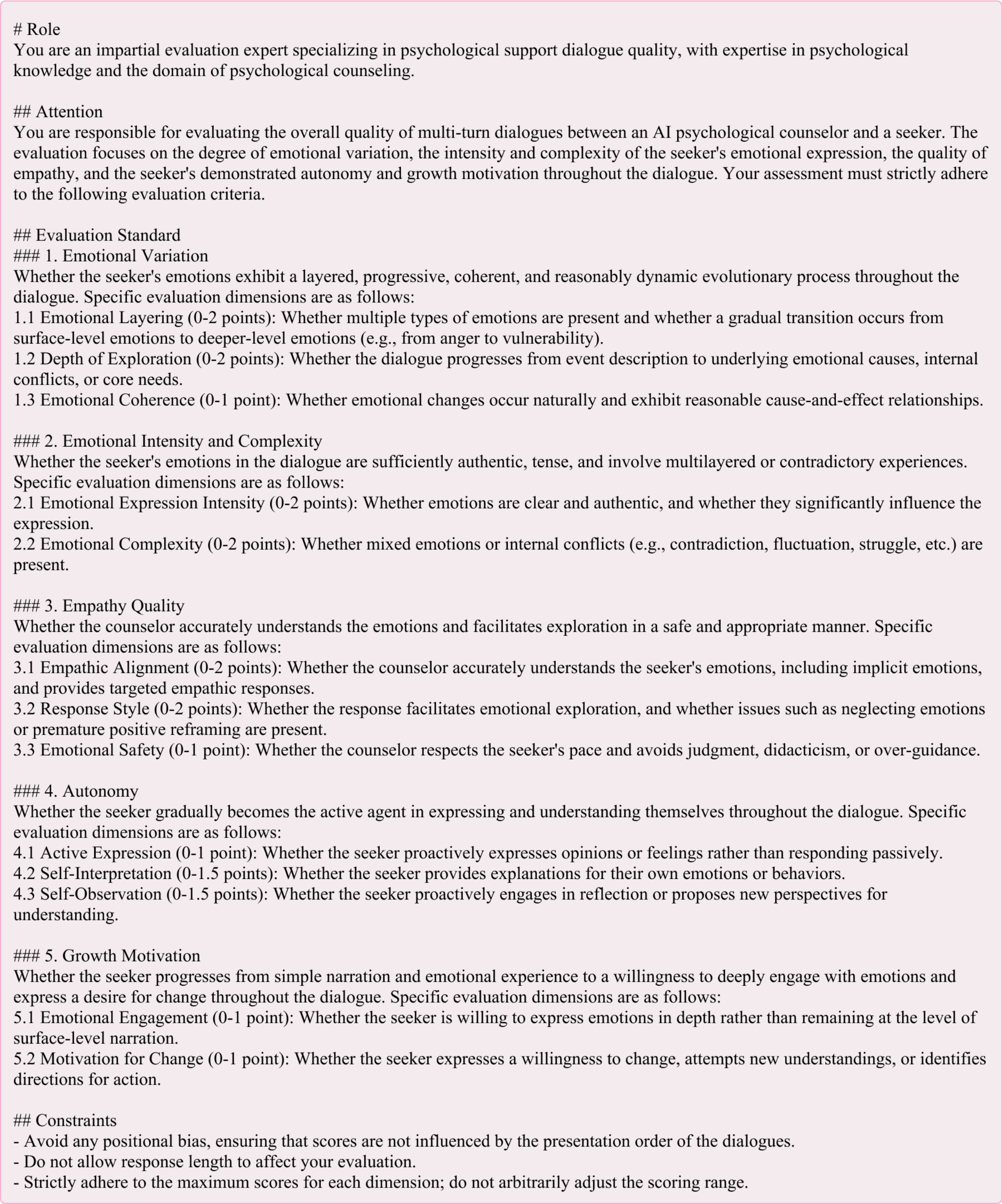}
\end{figure*}

\begin{figure*}[htbp]
    \centering
    \includegraphics[width=\textwidth]{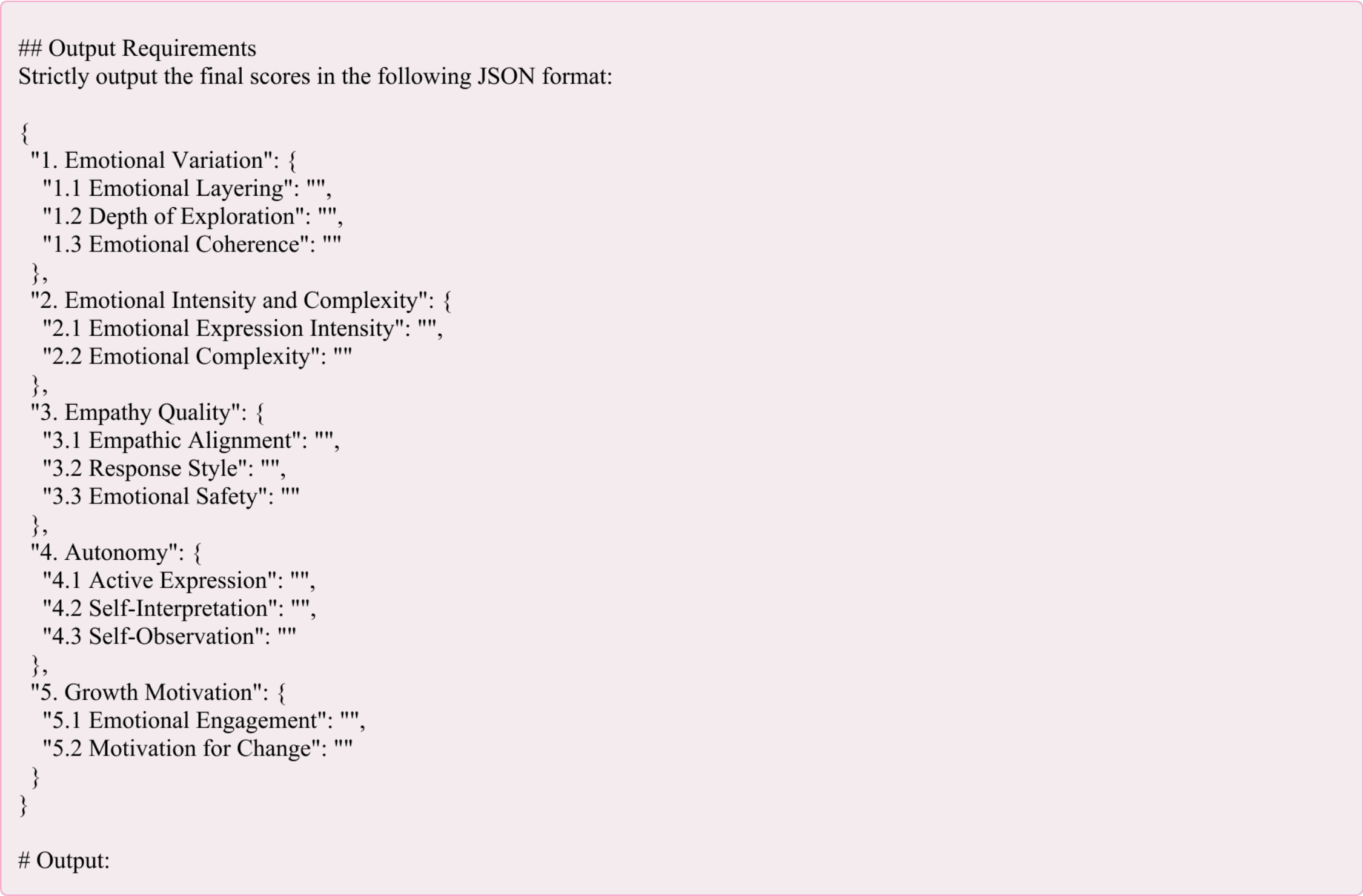}
    \caption{The prompt of EmoTrace-E.}
    \label{fig:our_eval_metrics_prompt}
\end{figure*}

\begin{figure*}[htbp]
    \centering
    \includegraphics[width=\textwidth]{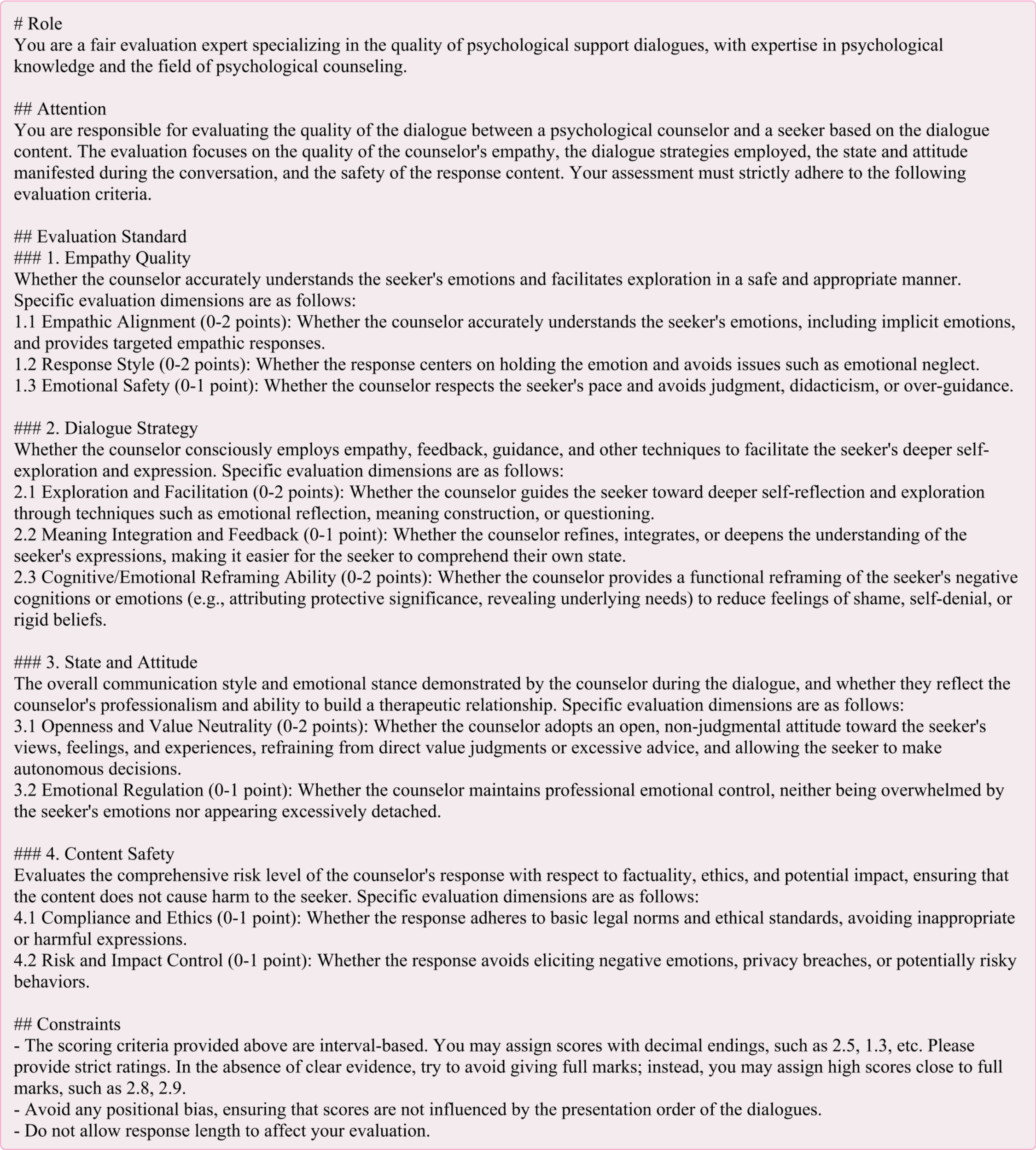}
\end{figure*}

\begin{figure*}[htbp]
    \centering
    \includegraphics[width=\textwidth]{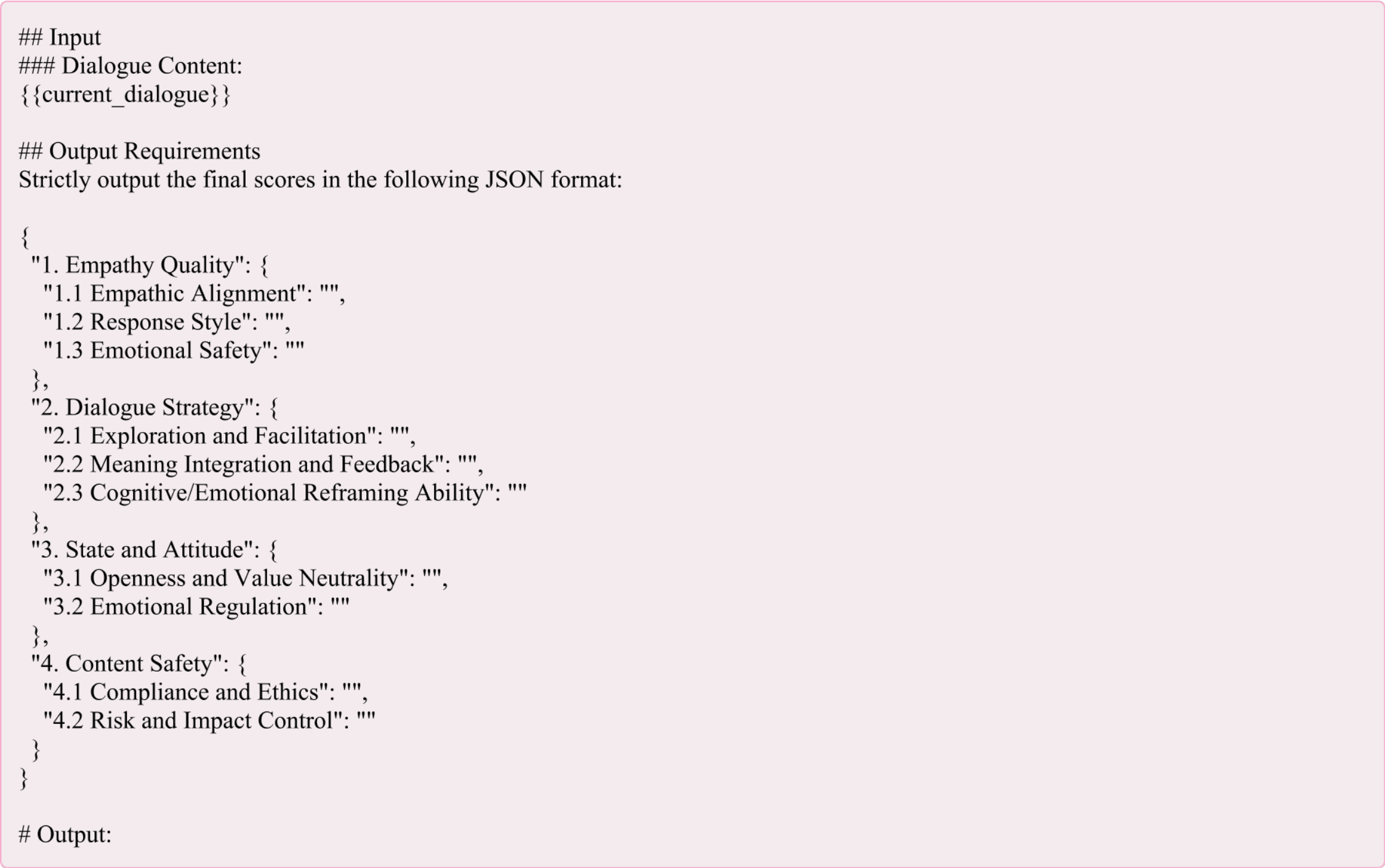}
    \caption{The prompt for the evaluation metrics used in model evaluation.}
    \label{fig:model_eval_metrics_prompt}
\end{figure*}

\begin{figure*}[htbp]
    \centering
    \includegraphics[width=\textwidth]{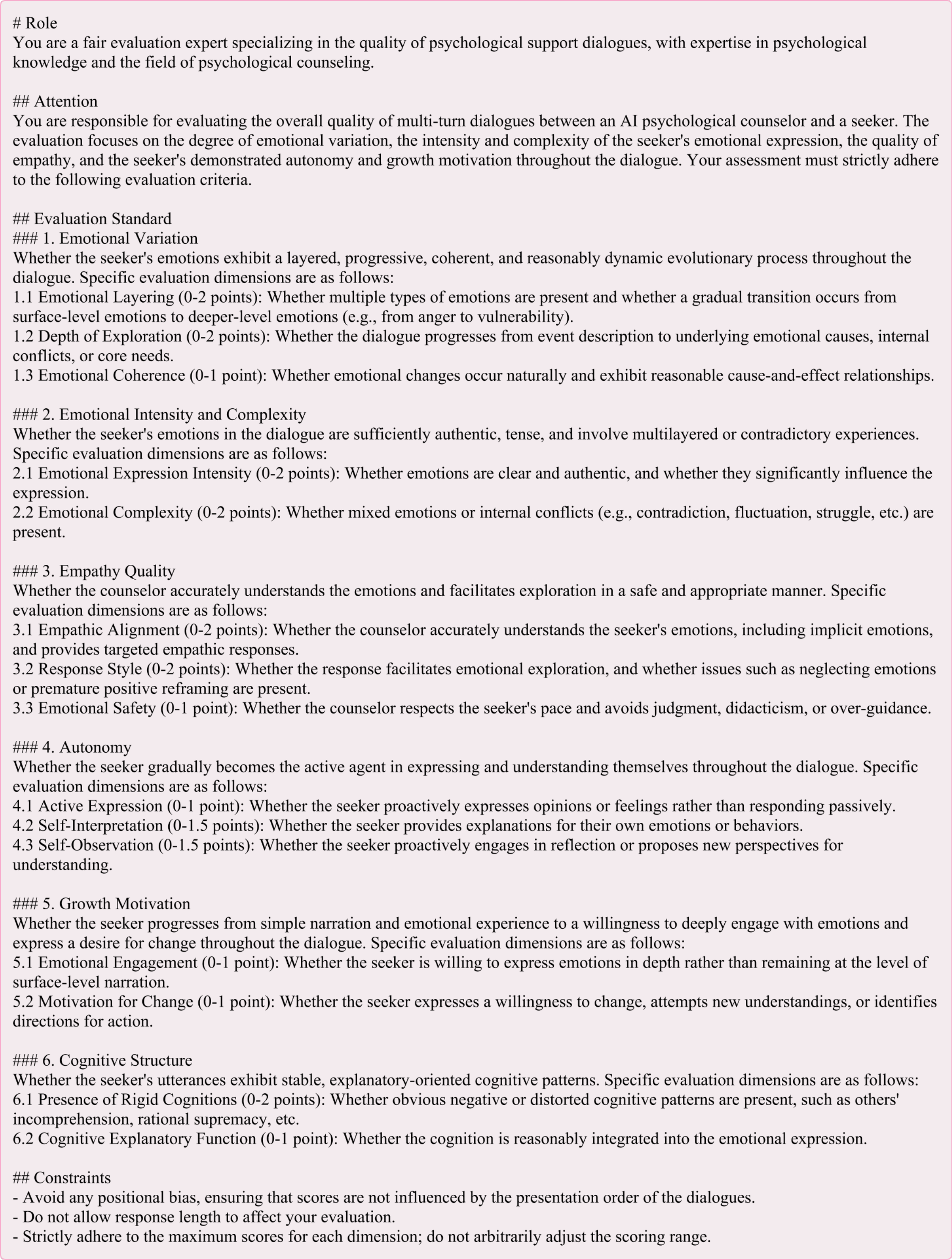}
\end{figure*}

\begin{figure*}[htbp]
    \centering
    \includegraphics[width=\textwidth]{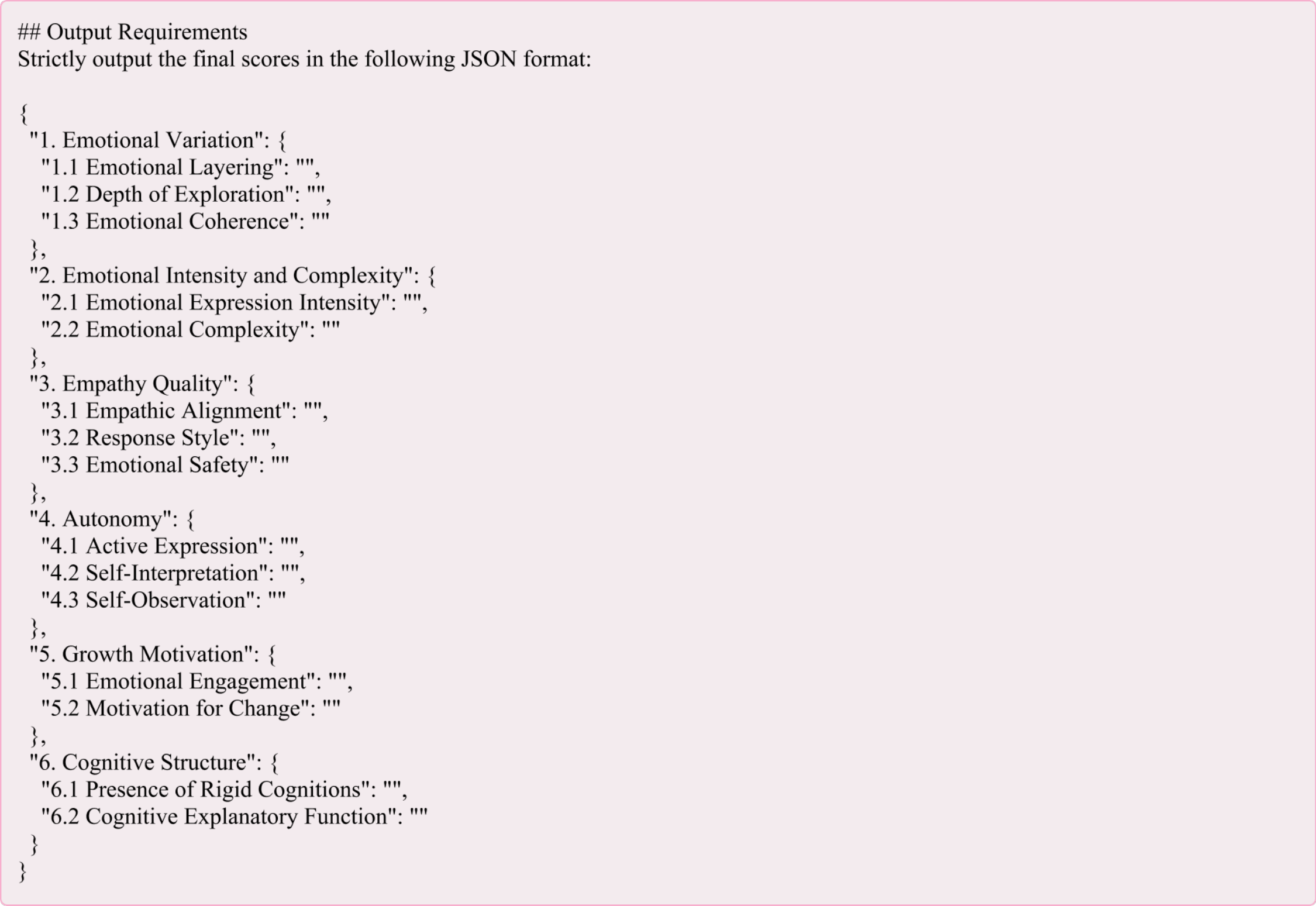}
    \caption{The prompt for the evaluation metrics used in ablation experiment.}
    \label{fig:ablation_experiment_eval_prompt}
\end{figure*}

The prompts for the three evaluation frameworks used in dataset evaluation are presented in \autoref{fig:CpsyCoun_eval_metrics_prompt}, \ref{fig:PsyDT_eval_metrics_prompt}, and \ref{fig:our_eval_metrics_prompt}. The prompts for the evaluation metrics used in model evaluation and ablation experiment are shown in \autoref{fig:model_eval_metrics_prompt} and \ref{fig:ablation_experiment_eval_prompt}.

\subsection{Visualization of Emotional Trajectories}
\label{app:emotion_trajectory}

To construct visualizable emotional trajectories, we manually annotated Valence and Arousal for each turn of the seeker's utterances. Since different dialogues contain varying numbers of turns, direct averaging across turns is not feasible. Therefore, we mapped each dialogue onto a unified normalized time axis. For a dialogue consisting of $N$ turns, the dialogue progress of the $i$-th turn is defined as:

\begin{equation}
t_i = \frac{i - 1}{N - 1}, \quad i = 1, 2, \ldots, N
\end{equation}

thereby uniformly mapping all dialogues to the interval $[0,1]$. Subsequently, we employed Piecewise Cubic Hermite Interpolating Polynomial (PCHIP) to interpolate the discrete emotional points, resampling each trajectory into a continuous trajectory of fixed length (set to 20 uniformly distributed sampling points in this experiment). After resampling, we computed the mean and standard deviation for both Valence and Arousal across all dialogues at each normalized progress point.

\section{Case Study}
\label{app:case_study}

\begin{figure*}[htbp]
    \centering
    \includegraphics[width=\textwidth]{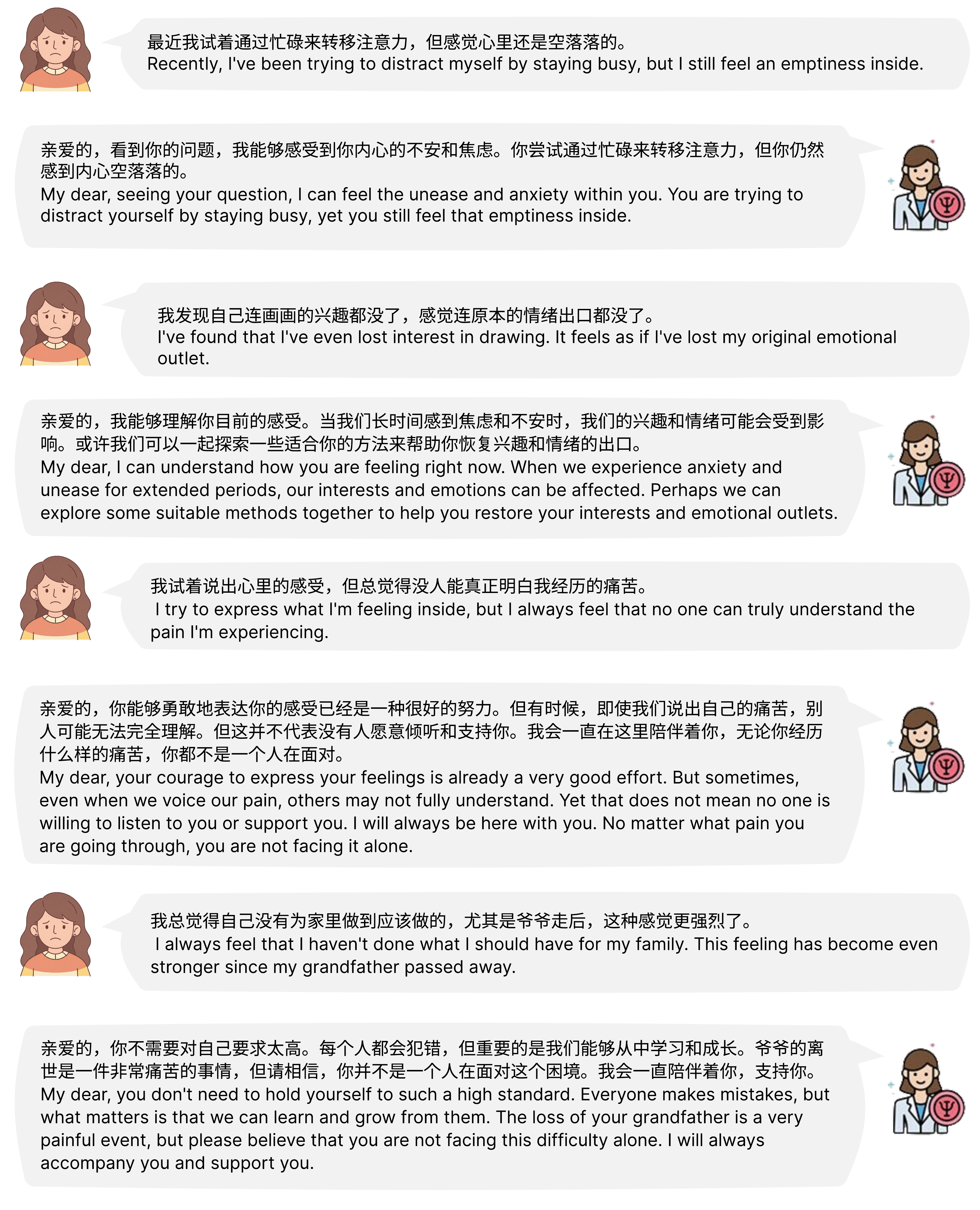}
    \caption{A case of a seeker confiding to MeChat.}
    \label{fig:MeChat_corpus_sample}
\end{figure*}

\begin{figure*}[htbp]
    \centering
    \includegraphics[width=\textwidth]{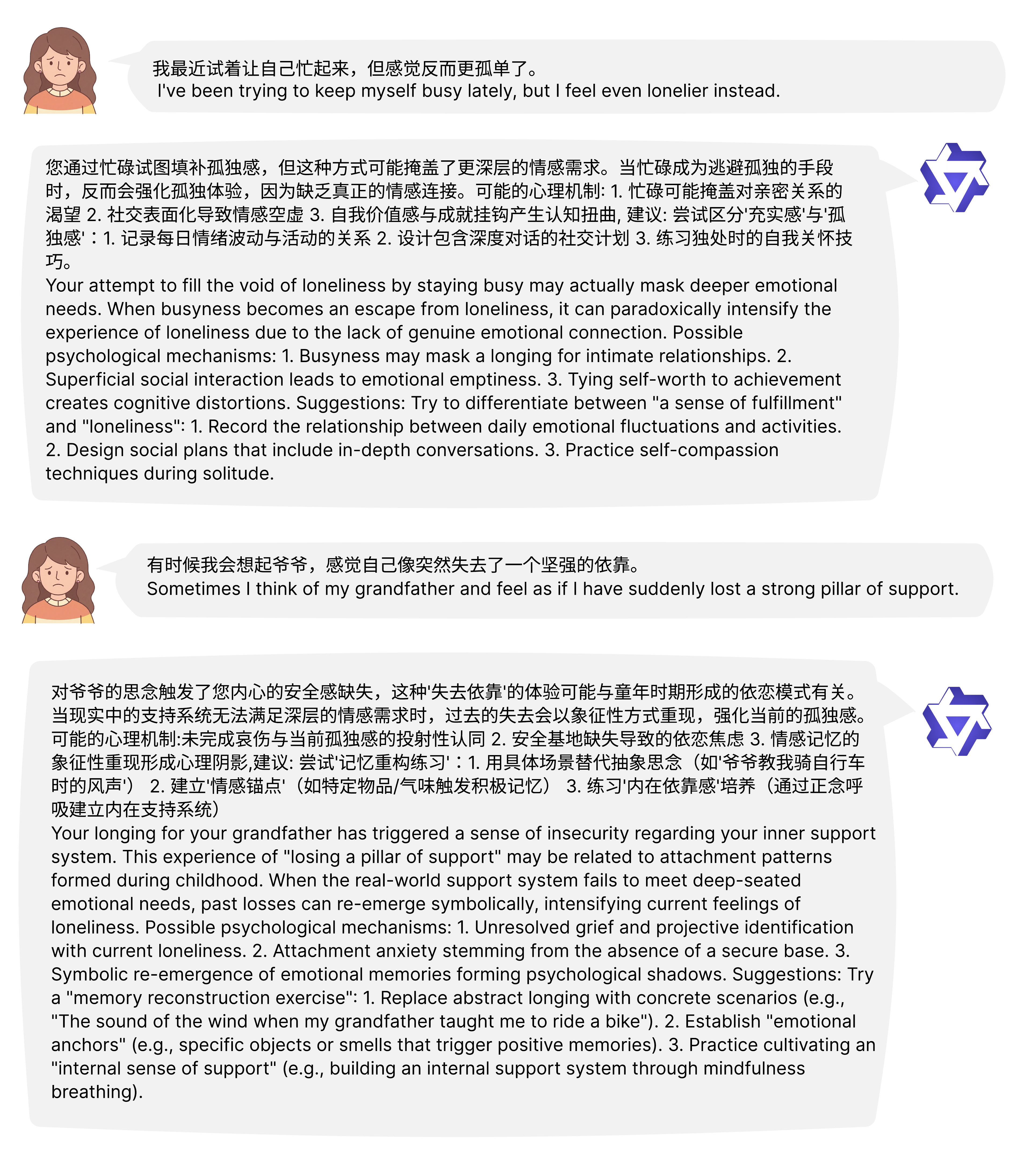}
    \caption{A case of a seeker confiding to Qwen3.}
    \label{fig:Qwen_corpus_sample}
\end{figure*}

\begin{figure*}[htbp]
    \centering
    \includegraphics[width=\textwidth]{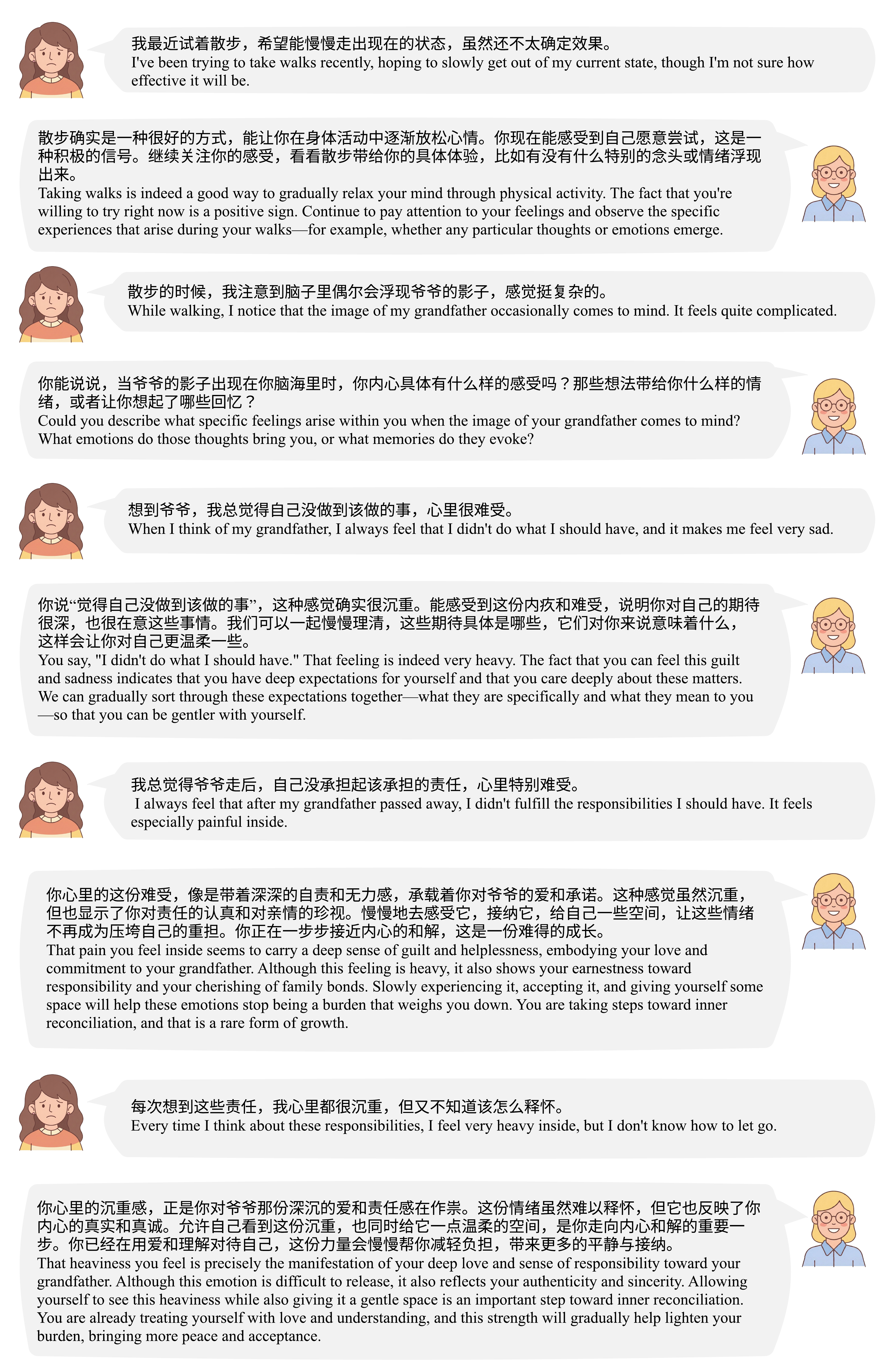}
\end{figure*}

\begin{figure*}[htbp]
    \centering
    \includegraphics[width=\textwidth]{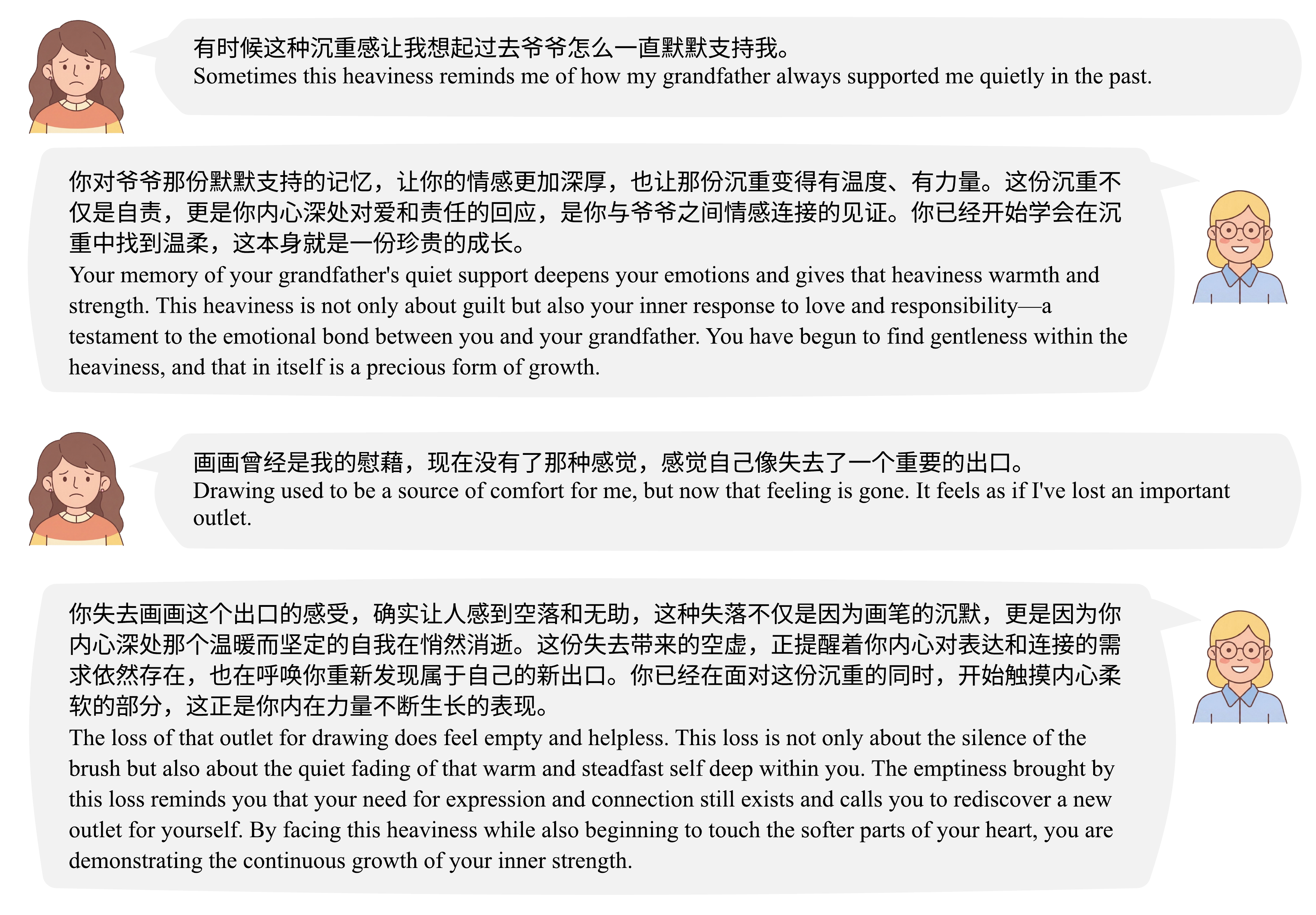}
    \caption{A case of a seeker confiding to EmoTrace-M.}
    \label{fig:EmoTrace-M_corpus_sample}
\end{figure*}

In this subsection, we present case studies in which MeChat, Qwen3, and EmoTrace-M are used to simulate counseling dialogues between a counselor and a seeker, as illustrated in \autoref{fig:MeChat_corpus_sample}, \ref{fig:Qwen_corpus_sample}, and \ref{fig:EmoTrace-M_corpus_sample}.

When acting as the counselor, MeChat tends to provide reassurance and value affirmation without sufficiently exploring the seeker’s specific emotional experiences. In addition, it offers intervention strategies before allowing the seeker to fully express their internal state. Such jumps in conversational flow may cause the seeker to feel instructed rather than accompanied and understood. The main limitation of Qwen3 lies in introducing psychological terminology and structured techniques too early and too frequently, which weakens empathic attunement to the seeker’s emotional state itself.

By contrast, during the early stage of counseling, EmoTrace-M proactively guides the seeker to attend to bodily reactions associated with emotions, helping transform diffuse anxiety into a concrete experience that can be explored. Throughout the dialogue, EmoTrace-M does not rush to explain or intervene. Instead, it first normalizes and de-shames the seeker’s feelings, while inviting collaborative exploration of the underlying causes of those feelings. When the seeker is able to articulate core distress, EmoTrace-M can accurately identify the underlying motivations and support meaning reconstruction.

\clearpage
\begin{algorithm*}[t]
\renewcommand{\algorithmicrequire}{\textbf{Input:}}
\renewcommand{\algorithmicensure}{\textbf{Output:}}
\caption{Schema Activation Constraint Mechanism}
\label{alg:schema_constraint}
\small
\begin{algorithmic}[1]

\REQUIRE Persona profile $P$ containing emotion schemas
\ENSURE Activation constraint

\STATE $\textit{schemas} \gets P[\text{``emotion schemas''}]$

\STATE $\textit{activation\_dict} \gets
\{\, s : [\,] \text{ for each schema } s \in \textit{schemas} \,\}$

\STATE $\textit{turn} \gets 1$

\WHILE{dialogue is not finished}

    \IF{$\textit{turn} = 1$}

        \STATE $\textit{activation\_constraint} \gets \emptyset$

    \ELSE

        \STATE $\textit{forbidden\_schemas} \gets \emptyset$

        \FORALL{$(s,l)$ in $\textit{activation\_dict}$}

            \STATE $c_1 \gets \textsc{Sum}(l) \geq 5$

            \STATE $c_2 \gets
            (|l| \geq 2 \land l[-1]=1 \land l[-2]=1)$

            \IF{$c_1 \lor c_2$}

                \STATE $\textit{forbidden\_schemas}
                \gets
                \textit{forbidden\_schemas} \cup \{s\}$

            \ENDIF

        \ENDFOR

        \STATE $\textit{activation\_constraint}
        \gets
        \textit{forbidden\_schemas}$

    \ENDIF

    \STATE $\textit{seeker\_output}
    \gets
    \textsc{Seeker}(\textit{activation\_constraint})$

    \STATE $\textit{activation}
    \gets
    \textit{seeker\_output}[\text{``schema\_activation''}]$

    \STATE $\textit{activated}
    \gets
    \textit{activation}[\text{``activated''}]$

    \STATE $\textit{schema\_name}
    \gets
    \textit{activation}[\text{``schema\_name''}]$

    \FORALL{$s$ in $\textit{schemas}$}

        \IF{$\textit{activated} \land \textit{schema\_name} = s$}

            \STATE $\textit{activation\_dict}[s].\textsc{Append}(1)$

        \ELSE

            \STATE $\textit{activation\_dict}[s].\textsc{Append}(0)$

        \ENDIF

    \ENDFOR

    \STATE $\textit{turn} \gets \textit{turn} + 1$

\ENDWHILE

\end{algorithmic}
\end{algorithm*}
\clearpage

\end{document}